\documentclass[times,twocolumn,final]{elsarticle}
\usepackage{Style/medima}
\usepackage{Style/slashbox}

\usepackage{framed,multirow}
\usepackage{bm}
\usepackage{amssymb}
\usepackage{latexsym}

\usepackage{url}
\usepackage{xcolor}

\usepackage {amsmath}
\usepackage {amssymb}
\usepackage {bm}
\usepackage {graphicx}
\usepackage {graphpap}
\usepackage {longtable}
\usepackage {multirow}
\usepackage {tikz}
\usetikzlibrary {decorations.markings}
\usepackage {varioref}
\usepackage{algorithm}  
\usepackage[noend]{algpseudocode} 
\usepackage{enumitem}
\newcommand{\norm}[1]{\|#1\|}

\usepackage{hyperref}
\definecolor{newcolor}{rgb}{.8,.349,.1}

\usepackage[compact]{titlesec}
\titlespacing{\section}{0pt}{2ex}{1ex}
\titlespacing{\subsection}{0pt}{1ex}{0ex}
\titlespacing{\subsubsection}{0pt}{0.5ex}{0ex}
\usepackage{booktabs}

\renewcommand{\cite}{\citep}
\journal{Medical Image Analysis}
\begin{document}
\verso{A. Goparaju \textit{et al.}}

\begin{frontmatter}

% \title{On the evaluation and validation of off-the-shelf statistical shape modeling tools \\ in clinical applications}
\title{Benchmarking off-the-shelf statistical shape modeling tools in clinical applications}

\author[1,2]{Anupama \snm{Goparaju}}
%\ead{anupama.goparaju@sci.utah.edu}
%  
\author[3]{Alexandre \snm{B$\hat{\text{o}}$ne}}%\fnref{fn1}}
%\ead{alexandre.bone@icm-institute.org}
%
\author[4]{Nan \snm{Hu}}
%\ead{Nan.Hu@hci.utah.edu}
%
\author[5]{Heath B. \snm{Henninger}}
%\ead{heath.henninger@utah.edu}
%
\author[1,5]{Andrew E. \snm{Anderson}}
% \ead{andrew.anderson@hsc.utah.edu}
%
%\author[1,2]{Ross T. \snm{Whitaker}}
%\ead{whitaker@sci.utah.edu}
%
\author[3]{Stanley \snm{Durrleman}}
%\ead{stanley.durrleman@inria.fr}
%
\author[5]{Matthijs \snm{Jacxsens}}
% \ead{matthijs.jacxsens@gmail.com}
%
\author[6]{Alan \snm{Morris}}
% \ead{Alan.Morris@carma.utah.edu}
%
\author[6]{Ibolya \snm{Csecs}}
% \ead{ibolyacsecs@gmail.com}
%
%\author[7]{Evgueni \snm {Kholmovski}}
%\ead{Evgueni.Kholmovski@hsc.utah.edu}
%
\author[6]{Nassir \snm{Marrouche}}
% \ead{nassir.marrouche@carma.utah.edu}
%
\author[1,2]{Shireen Y. \snm{Elhabian}\corref{cor1}}
\cortext[cor1]{Corresponding author: Shireen Elhabian \\ \textit{Address:} Scientific Computing and Imaging Institute \\
72 S Central Campus Drive, Room 2815, Salt Lake City, UT 84112}
\ead{shireen@sci.utah.edu}

\address[1]{Scientific Computing and Imaging Institute, University of Utah, Salt Lake City, UT, USA}
\address[2]{School of Computing, University of Utah, Salt Lake City, UT, USA}
\address[3]{ARAMIS Lab, ICM, Inserm U1127, CNRS UMR 7225, Sorbonne University, Inria, Paris, France}
\address[4]{Robert Stempel School of Public Health and Social Work, Florida International University, Miami, FL, USA}
\address[5]{Department of Orthopaedics, School of Medicine, University of Utah, Salt Lake City, UT, USA}
\address[6]{Division of Cardiovascular Medicine, School of Medicine, University of Utah, Salt Lake City, UT, USA}
% \address[6]{Section of Cardiology, Heart \& Vascular Institute School of Medicine, School of Medicine, Tulane University, New Orleans, LA, USA}
%\address[7]{Department of Physical Therapy, University of Utah, Salt Lake City, UT, USA}
%\address[8]{Department of Biomedical Engineering, University of Utah, Salt Lake City, UT, USA}

\received{\today}
\finalform{}
\accepted{}
\availableonline{}
\communicated{}

\begin{abstract}
%%%

%%The abstract should be no longer than 200 words. 
Statistical shape modeling (SSM) is widely used in biology and medicine as a new generation of morphometric approaches for the quantitative analysis of anatomical shapes.
Technological advancements of \textit{in vivo} imaging have led to the development of open-source computational tools that automate the modeling of anatomical shapes and their population-level variability.   
However, little work has been done on the evaluation and validation of such tools in clinical applications that rely on morphometric quantifications (e.g., implant design and lesion screening).
Here, we systematically assess the outcome of widely used, state-of-the-art SSM tools, namely ShapeWorks, Deformetrica, and SPHARM-PDM.
We use both quantitative and qualitative metrics to evaluate shape models from different tools. We propose validation frameworks for anatomical landmark/measurement inference and lesion screening. 
We also present a lesion screening method to objectively characterize subtle abnormal shape changes with respect to learned population-level statistics of controls.
Results demonstrate that SSM tools display different levels of consistencies, where ShapeWorks and Deformetrica models are more consistent compared to models from SPHARM-PDM due to the groupwise approach of estimating surface correspondences. 
Furthermore, ShapeWorks and Deformetrica shape models are found to capture clinically relevant population-level variability compared to SPHARM-PDM models.
%%%%
\end{abstract}

\begin{keyword}
\KWD \\
Statistical shape models \\
Population analysis \\
Correspondence optimization \\
Surface parameterization \\
Algorithm evaluation and validation \\
Landmark inference \\
Lesion screening 
\end{keyword}

\end{frontmatter}

% \linenumbers
% \newpage

%%% -*-LaTeX-*-

\section{Introduction}

%-------------------------------------------------------------------------------------
\textit{Shape} is the geometric information that remains when all the global geometrical properties are factored out, such as translation, orientation, and size (depending on the study at hand) \cite{mardia1989statistical}. 
Since the pioneering work of D'Arcy Thompson \cite{thomson1917growth}, \textit{morphometrics} (or \textit{shape analysis}) has evolved into an indispensable quantitative tool in medical and biological sciences to study shapes. 
Shape analysis has several applications in archaeology \cite{woods2017potential}, medical imaging \cite{joskowicz2018future,heimann2009statistical}, computer-aided design \cite{joskowicz2018future,zadpoor2015patient,kozic2010optimisation}, and biomechanics \cite{bredbenner2014development,nicolella2012development}.
% surgical planning \cite{markelj2012review,zheng20092d}, evolutionary biology \cite{dominguez2012utility}, genetics \cite{twigg2009skeletal}, and cancer research  \cite{liu2015female,cates2017shape}. 

%-------------------------------------------------------------------------------------

\textit{Statistical shape modeling} (SSM) is the application of mathematics, statistics, and computing to parse the shape into some quantitative representation that will facilitate testing of biologically relevant hypotheses.
SSM can help answer various questions about the population under study. Among the many examples, SSM can answer whether a specific bone can be used to classify a group of species in evolutionary biology \cite{dominguez2012utility}, how a gene mutation contributes to skeletal development \cite{twigg2009skeletal}, the shape changes of brain structures in patients with depression and schizophrenia \cite{styner2006framework,zhao2008hippocampus,davies2003shape}, and the extent of bone deformation due to genetic diseases that can cause a specific type of cancer \cite{liu2015female,cates2017shape}.
The quantitative, population-level analysis of anatomical shapes can also assist in different \textit{clinical applications}, including disease diagnosis \cite{shinya2011}, optimal implant design and selection \cite{goparaju2018evaluation}, anatomy reconstruction and segmentation \cite{gollmer2014} from medical images for computer-aided surgery \cite{zachow2015computational}, and preoperative and postoperative surgical planning \cite{rodriguez2017statistical,markelj2012review,zheng20092d}. %For instance, shape abnormalities compared to the control population can be an indication of pathology. 
These advancements in biomedical and clinical applications that benefit from SSM have the potential to make clinical-decision making more objective.

%-------------------------------------------------------------------------------------

Computational tools for shape modeling define an \textit{anatomical mapping}, i.e., \textit{metric}, among shapes to enable quantifying subtle shape differences (i.e., comparing shapes) and performing shape statistics (e.g., averaging). That is, the shapes that differ in a manner that is typical of the shape variability (e.g., size is a typical anatomical variation) in the population are considered similar compared to the shapes that differ in atypical ways. For example, extra bone growth on a femur (a bone that comprises the distal segment of the hip joint) that is indicative of a pathology differs from a control femur in an atypical way. 
A growing consensus in the field is that such a metric should be adapted to the specific population under investigation, which entails finding \textit{correspondences} across an ensemble of shapes \cite{srivastava2005statistical,kulis2013metric}.
Manually defined landmarks, defined consistently on each shape instance (i.e., homology), have been the most popular choice for a light-weight shape representation that is suitable for statistical analysis and visual communication of the results \cite{zachow2015computational,sarkalkan2014statistical}.
However, manual annotation is tedious, time-consuming, expert-driven (and hence subjective), and even prohibitive for three-dimensional (3D) shapes, especially with large shape ensembles. 
SSM is an important shift from manually defined anatomical homologies to computationally derived (i.e., automated) correspondence (shape) models.
Finding correspondences across an ensemble of shapes can be posed as an optimization problem leading to the development of various open-source SSM tools. 

% The scientific premise of existing correspondence techniques falls in two broad categories: a \textit{groupwise} approach to estimating 
% correspondences (e.g., ShapeWorks \cite{cates2017shapeworks}, Minimum Description Length - MDL \cite{davies2002learning}, Deformetrica \cite{durrleman2014morphometry}) that considers the variability in the entire cohort and a \textit{pairwise} approach (e.g., SPHARM-PDM \cite{styner2006framework}) that considers mapping to a predefined surface parameterization. 
% %
% Pairwise methods lead to biased and suboptimal models \cite{oguz2015entropy,dalal2010multiple,davies2008statistical}. On the other hand, groupwise methods learn a population-specific metric in a way that does not penalize natural variability and therefore can capture the underlying parameters in an anatomical shape space. 

The scientific premise of existing correspondence techniques falls in two broad categories, pairwise and groupwise \cite{oguz2015entropy}. The \textit{pairwise} approach treats each shape instance independently and estimates correspondences by mapping the subject to a predefined atlas or template (e.g., SPHARM-PDM \cite{styner2006framework}). The \textit{groupwise} approach, on the other hand, estimates point correspondences by considering the variability in the entire cohort of shapes to quantify the quality of correspondences (e.g., ShapeWorks \cite{cates2017shapeworks}, Minimum Description Length - MDL \cite{davies2002learning}, Deformetrica \cite{durrleman2014morphometry}).
%
% Pairwise methods lead to biased and suboptimal models \cite{oguz2015entropy,dalal2010multiple,davies2008statistical}. On the other hand, 
Hence, groupwise methods learn a population-specific metric in a way that does not penalize natural variability and therefore can capture the underlying parameters in an anatomical shape space. 
%
% These SSM approaches can result in different shape models (see Figure \ref{BoxBump} for a simple example), which could directly affect the performance of the resulting shape model in real-world clinical applications. 
%
Other publicly available tools, e.g., FreeSurfer \cite{fischl1999high}, BrainVoyager \cite{goebel2006analysis}, FSL \cite{jenkinson2012fsl}, and SPM \cite{ashburner2012spm}, provide shape modeling capabilities, but they are tailored to specific anatomies or limited topologies. 
Shape \textit{analysis} tools, such as R shapes package \cite{Dryden-shapes} and MorphoJ \cite{klingenberg2011morphoj}, require point correspondences (defined manually or automatically via an SSM tool) for the input shapes to perform statistical analysis.

%-------------------------------------------------------------------------------------

Better understanding of the consequences of different SSM tools for the final analysis is critical for the careful choice of the tool to be deployed for a clinical application. 
This study is thus motivated by the potential role of SSM in clinical scenarios that (1) are driven by anatomical measurements, which could be automated by relating patient-level anatomy to population-level morphometrics, and (2) entail pathology (lesion) screening, which could be informed by population-level statistics of controls.
In this paper, we significantly extend the preliminary analysis presented in \cite{goparaju2018evaluation} to expand the clinical application under analysis.
In particular, we demonstrate the significance of evaluation and validation of SSM tools in the context of clinical applications, such as implant design and selection, motion tracking, surgical planning, and screening of bony lesions.
Here, we consider a representative set of open-source, widely used SSM tools that support shape modeling of general anatomies; namely ShapeWorks \cite{cates2017shapeworks}, Deformetrica \cite{durrleman2014morphometry}, and SPHARM-PDM \cite{styner2006framework} (recently incorporated into SlicerSALT \cite{vicory2018slicersalt}).
%
% Applications of ShapeWorks include identification of group differences in the mean control and mean pathologic unstable scapulae \cite{jacxsens2019coracoacromial}, detection of variations in femoral morphology due to femoroacetabular impingement (FAI) \cite{harris2013statistical}, and use of left atrial appendage shape as a predictor of atrial fibrillation recurrence \cite{bieging2018left}. 
% %
% Applications of Deformetrica include quantitative assessment of craniofacial surgery \cite{rodriguezquantitative}, classification of patients with Alzheimer’s disease \cite{routier2014evaluation}, and cranioplasty surgical planning \cite{rodriguez2017statistical}.
% %
% Applications of SPHARM-PDM include boundary and medical shape analysis of the hippocampus in schizophrenia \cite{styner2004boundary}, orthognathic surgical displacement analysis \cite{paniagua2011clinical}, and quantification of temporomandibular joint osteoarthritis. 
%
We propose evaluation and validation frameworks for anatomical landmark/measurement inference and lesion screening. 
We also present a lesion screening method to provide an objective characterization of subtle abnormal shape changes with respect to learned population-level statistics of controls.

\section {Related work}

%-------------------------------------------------------------------------------------

Open-source SSM tools rely on different modeling approaches and assumptions to establish surface correspondences.
However, evaluating shape models is a nontrivial task due to the lack of ground-truth correspondences. 
Shape models can be \textit{intrinsically} evaluated using quantitative metrics that reflect the correspondence quality \cite{davies2002learning}.
% the correspondence quality via metrics such as \textit{compactness}, the percentage of variance captured by a specific number of modes; \textit{generalization}, the ability to represent unseen shape instances; and \textit{specificity}, the ability to generate plausible shapes \cite{davies2002learning}.
%
However, such metrics have been criticized since relevant shape information may be lost while still obtaining excellent evaluation measures \cite{ericsson2007measures}. 
% assessing the correspondence solely based on these metrics has been criticized since relevant shape information may be lost while still obtaining excellent evaluation measures \cite{ericsson2007measures}. 
% excellent measures of specificity, generality and compactness \cite{ericsson2007measures}. 
% For instance, perfect compactness can be obtained by representing all shapes in the population by a single landmark placed in the same spatial location \cite{ericsson2007measures}. 
Hence, there is an unmet need to benchmark SSM tools via \textit{extrinsic} validation metrics that signify the impact of shape models in clinical applications. 

%-------------------------------------------------------------------------------------

%Ericsson and Karlsson 
\cite{ericsson2007measures} relied on manually picked landmarks to validate the computationally derived correspondences. 
%
%Munshell et al. 
\cite{munsell2008evaluating} developed a similar approach to benchmark correspondence optimization techniques using synthetic shapes.
%with the ability to recover the ground-truth shape model, given a synthetic set of shapes. 
%
These two approaches require ground-truth correspondences to evaluate shape models, which is not trivial, and is instead prohibitive, for nonsynthetic 3D shapes (e.g., anatomies). 
%
%Munshell et al. 
Furthermore, \cite{munsell2008evaluating} conducted experiments for 2D shapes. Extending these techniques to 3D shapes would not be feasible, given the complexity of medical and biological shapes. These evaluation studies have theoretical grounds, yet have not considered real-world applications. 

%-------------------------------------------------------------------------------------

Very few studies have evaluated SSM tools in the context of biomedical applications. SSM tools have been evaluated in nonclinical applications such as image segmentation to quantify the influence of a shape model on the image segmentation accuracy \cite{gollmer2014}.
\cite{gao2014} proposed a framework for the generation of synthetic, ground-truth correspondences via a shape-deformation synthesis approach to compare shape models from SPHARM-MAT, SPHARM-PDM, ShapeWorks, and tensor-based morphometry (TBM). This study focused on shapes with simple geometric complexity (e.g., caudate) and simulated pathologies. The comparison of the shape models found inconsistencies and disagreement among the  different tools. 
However, little work has been done in the evaluation and validation of SSM tools in clinical applications. Hence, a systematic evaluation and validation framework that enables assessment of shape models from different tools can assist in SSM tool selection in clinical scenarios.

%-------------------------------------------------------------------------------------

\begin{figure}%[tb!]
    \centerline{\includegraphics[width = 1\linewidth]{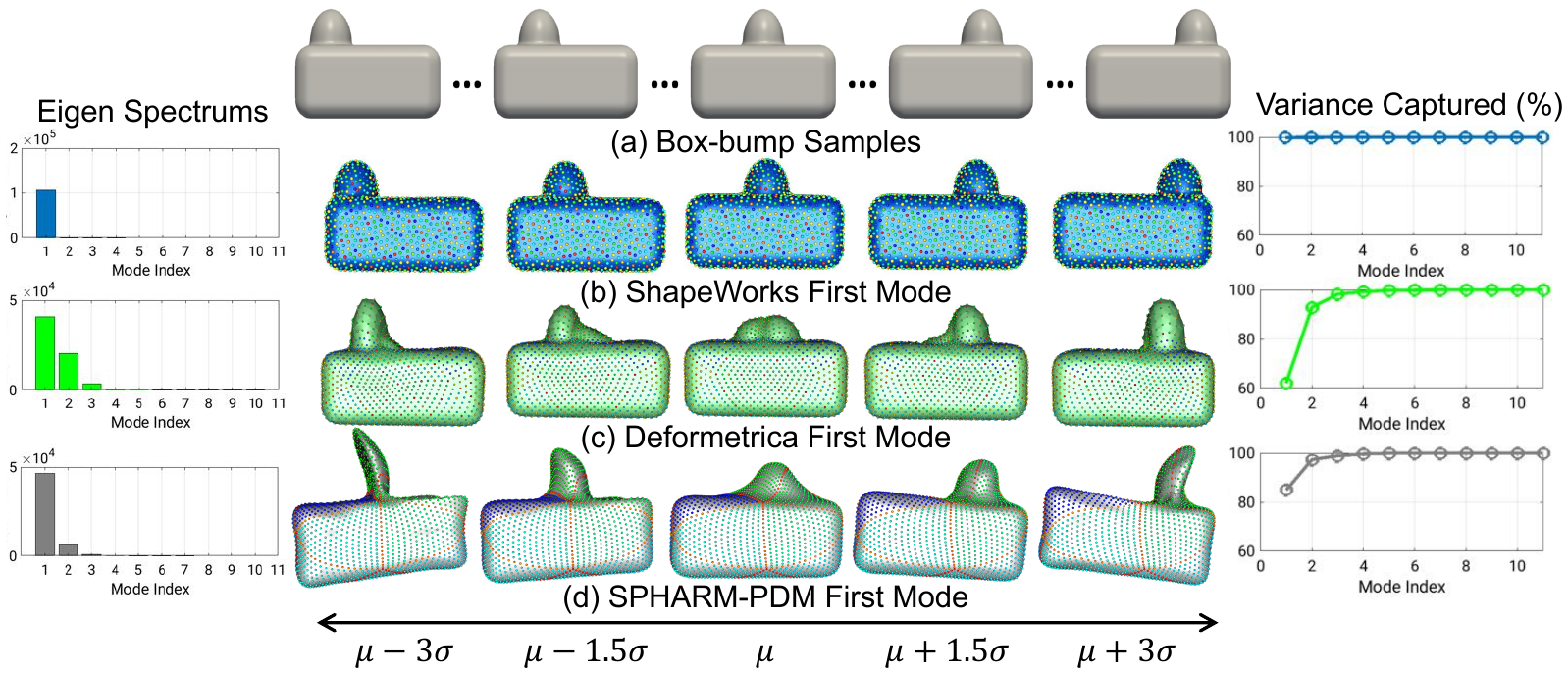}}
    \caption{Mode of variation. (a) Box-bump samples. The mean $\pm$ 3 standard deviations of the first dominant mode
of (b) ShapeWorks \cite{cates2017shapeworks}, (c) Deformetrica \cite{durrleman2014morphometry}, and (d) SPHARM-PDM \cite{styner2006framework}. Adapted with permission from \cite{goparaju2018evaluation}.}%
    \label{BoxBump}
\end{figure}

To demonstrate the need for and significance of SSM tool assessment, we performed a proof-of-concept experiment on an ensemble of 3D shapes of boxes with a moving bump, where computationally derived point correspondences were obtained using ShapeWorks \cite{cates2017shapeworks}, Deformetrica \cite{durrleman2014morphometry}, and SPHARM-PDM \cite{styner2006framework}. 
This example is interesting because we would, in principle, expect an SSM tool to discover a single mode of variability (i.e., the moving bump) by generating surface correspondences that respect the natural shape variability in the population. However, different tools have yielded different results (\figurename~\ref{BoxBump}). ShapeWorks \cite{cates2017shapeworks}, which adopts a groupwise approach, correctly discovered the underlying population variability and generated shape more faithful to those described by the training set, even out to three standard deviations. This proof-of-concept motivates the need to perform a systematic evaluation and validation of these SSM tools as related to application-specific clinical needs.

%-------------------------------------------------------------------------------------

\section{Background}
%-------------------------------------------------------------------------------------

Here, we give an overview of the SSM tools considered for the performance analysis and provide the clinical scenarios that can benefit from such analysis.

%-------------------------------------------------------------------------------------
\subsection{Statistical shape modeling (SSM) tools}\label{ssm-tools}
%-------------------------------------------------------------------------------------

A \textit{shape model} provides both a detailed 3D geometrical representation of the average anatomy of a given population and a representation of the population-level geometric variability of the anatomy, in the form of a collection of principal modes of variation.
SSM tools for point-based models automate the point-correspondence estimation of an ensemble of shapes via an \textit{optimization} problem that quantifies the notion of correspondences. Once correspondences are obtained (in a common coordinate system where rigid or similarity transformations are factored out), principal component analysis (PCA) can be performed to identify the dominant modes of variation in the shape space. Here, we overview the shape modeling approach pertaining to each of the considered SSM tools.

\begin{figure*}%[ptb!]
    \centerline{\includegraphics[width = 1\linewidth]{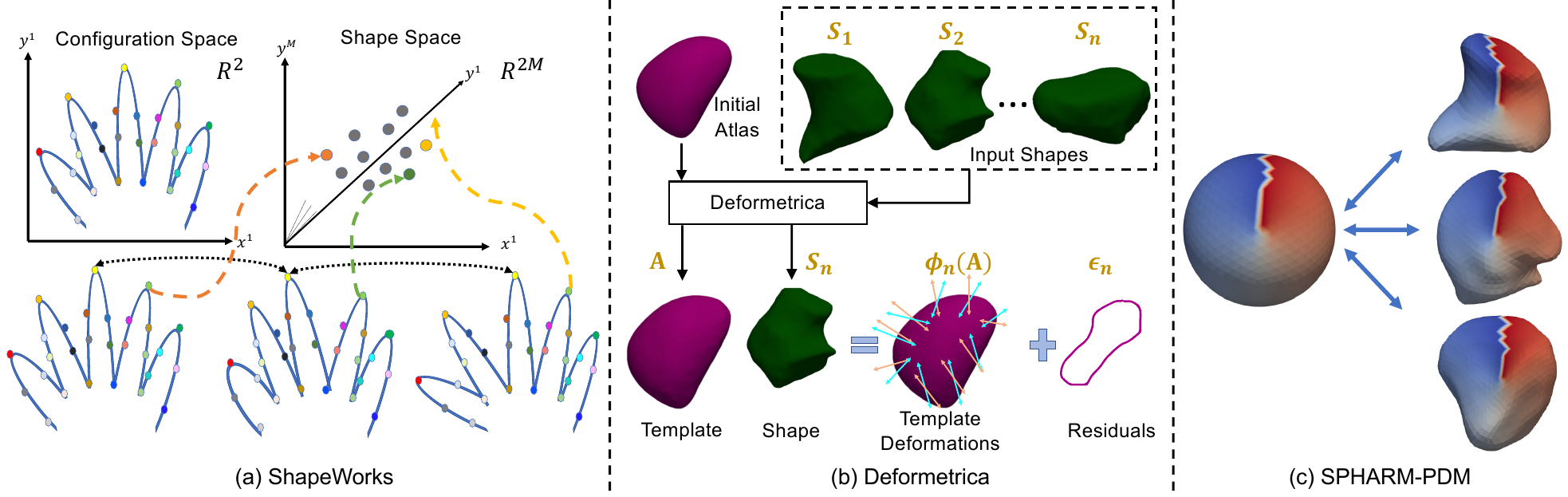}}
    \caption{SSM tools: (a) Shapeworks \cite{cates2017shapeworks} considers two random variables defining the configuration space and shape space. The configuration is a collection of $M$ point correspondences on a shape, which is mapped to a single point in the $dM-$dimensional shape space; 
    (b) Deformetrica \cite{durrleman2014morphometry} estimates a template from the set of input shapes and an initial atlas by generating point correspondences on the input shapes based on deformations; and (c) SPHARM-PDM \cite{styner2006framework} maps each input shape to a unit sphere through an area-preserving and distortion-minimizing objective using spherical harmonic basis functions, where color indicates correspondences between the sphere and the individual samples.}%
    \label{Tools}
\end{figure*}

%-------------------------------------------------------------------------------------
\subsubsection{ShapeWorks}
%-------------------------------------------------------------------------------------

ShapeWorks is a groupwise particle-based shape modeling (PSM) method \cite{cates2017shapeworks, cates2007shape} that is not constrained to any specific topology, handles open surfaces, and does not rely on any surface parameterizations.
The scientific and clinical utility of ShapeWorks has been demonstrated in a range of applications, including neuroscience \cite{datar2013geodesic,oguz2009cortical}, biological phenotyping \cite{jones2013toward,cates2017shape}, orthopaedics \cite{harris2013statistical,jacxsens2019coracoacromial,atkins2017quantitative}, and cardiology \cite{bieging2018left,bieging2018quantitative}.
% Applications of ShapeWorks include identification of group differences in the mean control and mean pathologic unstable scapulae \cite{jacxsens2019coracoacromial}, detection of variations in femoral morphology due to femoroacetabular impingement (FAI) \cite{harris2013statistical}, and use of left atrial appendage shape as a predictor of atrial fibrillation recurrence \cite{bieging2018left}. 
%
PSM formulation treats each surface as a collection of interacting dynamic \textit{particles} with mutually repelling forces to optimally cover, and therefore describe, the surface geometry. 
The correspondences are freely moving particles, yet they are constrained to lie on the surface, and their positions can be directly optimized.
This particle-based representation avoids many of the problems inherent in parametric representations such as the limitation to specific topologies, processing steps necessary to construct parameterizations, and bias toward model initialization using initial atlases.
%
%This formulation avoids the need to construct parameterizations and overcomes the limitations pertaining to specific topology and bias for the initialization. 

PSM optimization can be summarized as follows: Consider an ensemble of $N$ shapes $\mathcal{S}=$ \{$ \mathbf {x}_1, \mathbf {x}_2, ...\mathbf {x}_N$\}, each with its own set of $M$ particles (i.e., correspondences) $\mathbf {x}_n = [\mathbf {x}_n ^{1},\- \mathbf {x}_n ^{2}, \-...\mathbf {x}_n ^{M}]$, where ordering implies correspondence among shapes. A correspondence lives in a $d-$dimensional space, i.e., $\mathbf{x}_n ^{m} \in \mathbb{R} ^{d}$, with $d=2$ and $3$ for 2D and 3D shapes, respectively. %shapes and $d=3$ for 3D shapes.
For groupwise modeling, a rigid or similarity transformation $\mathbf{T}_n$ is estimated to transform the particles in the $n-$th shape \textit{local} coordinate system $\mathbf{x}_n ^{m}$ to the common coordinate system $\mathbf{z}_n ^{m}$ such that $\mathbf {z}_n ^{M}=\mathbf{T}_n \mathbf{x}_n ^{M}$.
This representation involves two types of random variables (\figurename~\ref{Tools}(a)): a \textit{shape space} variable $\mathbf{Z} \in \mathbb{R} ^{dM}$ and a particle position variable $\mathbf{X}_n \in \mathbb{R}^d$ that encodes particles distribution on the $n-$th shape (\textit{configuration space}).
Correspondences are optimized by minimizing a combined shape correspondence and surface sampling objective function $Q = H(\mathbf{Z}) - \sum_{n=1}^{N} H(\mathbf{X}_n)$, where $H$ is an entropy estimation assuming Gaussian shape distribution in the shape space and Euclidean particle-to-particle repulsion in the configuration space.
This formulation favors a compact ensemble representation in shape space (first term) against a uniform distribution of particles on each surface for accurate shape representation (second term).

%-------------------------------------------------------------------------------------
\subsubsection{Deformetrica}
%-------------------------------------------------------------------------------------

Deformetrica is a groupwise correspondence method that is based on the large deformation diffeomorphic metric mapping (LDDMM) framework \cite{durrleman2014morphometry}. This SSM tool is not constrained to any specific topology and supports open surfaces, but it requires an initial atlas that defines the topology of the shape class under study to estimate the template complex (i.e., average shape). 
Correspondences are not explicitly optimized; rather diffeomorphic deformations enable the correspondence establishment between the template complex and each input shape. The template complex captures the common characteristics of the shapes, and the deformations capture the variability in the shapes, as shown in \figurename~\ref{Tools}(b). 
Applications of Deformetrica include quantitative assessment of craniofacial surgery \cite{rodriguezquantitative}, classification of patients with Alzheimer’s disease \cite{routier2014evaluation}, and cranioplasty surgical planning \cite{rodriguez2017statistical}.

The diffeomorphic multiobject template complex construction is performed using a Bayesian framework \cite{gori2017bayesian}. The complex of a shape instance is modeled as a deformed template complex and a residual. The $n-$th shape complex is defined as $\mathbf{S}_{n} = \boldsymbol{\phi}_{n}(\mathbf{A}) + \boldsymbol{\epsilon} _{n}$, where $\boldsymbol{\phi}_{n}(\mathbf{A})$ is the deformation on the template ($\mathbf{A}$) specific to the $n-$th shape instance, and $\boldsymbol{\epsilon} _{n}$ is the residual. The variations in the shapes are modeled by these deformations, and each deformation is characterized by a set of parameters $\boldsymbol{\alpha} _{n}$. The assumption here is that the parameters follow a Gaussian distribution, with a mean 0 and a covariance matrix $\boldsymbol{\Gamma} _{\boldsymbol{\alpha}}$. The objective function is defined as estimating the template complex and covariance matrix by maximizing the joint posterior distribution of the shape complexes, i.e., 
$\{ \mathbf{A}^* , \boldsymbol{\Gamma} ^* _{\boldsymbol{\alpha}}\} =\operatorname*{argmin}_{\mathbf{T}, \boldsymbol{\Gamma} _{\boldsymbol{\alpha}}} p(\mathbf{A}, \boldsymbol{\Gamma} _{\boldsymbol{\alpha}} \lvert \{\mathbf{S}_n\}_{n=1}^{N})$. The maximization process is constrained by the requirement that the template complex should deform to match the shape complex, and the residual $\boldsymbol{\epsilon}_n$ should be small.

%-------------------------------------------------------------------------------------
\subsubsection{SPHARM-PDM}
%-------------------------------------------------------------------------------------

SPHARM-PDM is a pairwise parameterization-based correspondence method \cite{styner2006framework} that is restricted to anatomies with spherical topologies. The spherical parameterization is obtained by mapping each shape to a unit sphere through an area-preserving and distortion-minimizing objective using spherical harmonic (SPHARM) basis functions, as shown in \figurename~\ref{Tools}(c). The SPHARM description is obtained from the surface mesh and its spherical parameterization, which are then aligned using a first-order ellipsoid from the SPHARM coefficients to establish correspondences across shapes.
Applications of SPHARM-PDM include boundary and medical shape analysis of the hippocampus in schizophrenia \cite{styner2004boundary}, orthognathic surgical displacement analysis \cite{paniagua2011clinical}, and quantification of temporomandibular joint osteoarthritis. 

SPHARM basis functions ${Y_l}^k$ are defined with degree $l$ and order $k$, ${Y_l}^k(\theta, \phi) \- = \- \sqrt{\frac{2l+1}{4\pi} \frac{(l-k)!}{(l+k)!}}{P_l}^k\left(\cos\theta\right)e^{ik\phi}$, where $\theta \in [0;\pi]$, $\phi \in [0;2\pi]$, and ${P_l}^k$ the associated Legendre polynomials. 
The surface of the $n-$th shape can be expressed using SPHARM basis functions by decomposing three coordinate functions that define the surface as $\mathbf{x}_n(\theta, \phi)=\left(x_n(\theta, \phi), y_n(\theta, \phi), z_n(\theta, \phi)\right)^T$, and the surface would be of the form $\mathbf{x}_n(\theta, \phi)=\sum_{l=0}^{\infty}\sum_{k=-l}^{l} \mathbf{c}_n^{l,k} {Y_l}^k(\theta, \phi) $, where $\mathbf{c}_n^{l,k}$ are 3D coefficient vectors due to the three coordinate functions. These coefficients are obtained using a least-squares method to fit the $n-$th shape surface. A correspondence point $\mathbf{x}_n^m$ on the surface is given by a parameter vector ($\theta_m$, $\phi_m$), which represents the $m-$th location on the predefined sphere parameterization.

%-------------------------------------------------------------------------------------
\subsection{Clinical applications}\label{clinical-apps}
%-------------------------------------------------------------------------------------

Clinical applications, such as implant design and selection, surgical planning, bone resection, and bone grafting, require patient-specific anatomical representation, which can be automatically estimated by relating patient-specific anatomical shape to the learned population-level morphometrics. Such automation reduces manual and subjective clinical decisions \cite{shinya2011, rodriguez2017statistical}. In this paper, we consider a representative set of clinical needs that would benefit from SSM-informed decisions. 

\begin{figure}%[t!]
    \centerline{\includegraphics[width = 1\linewidth]{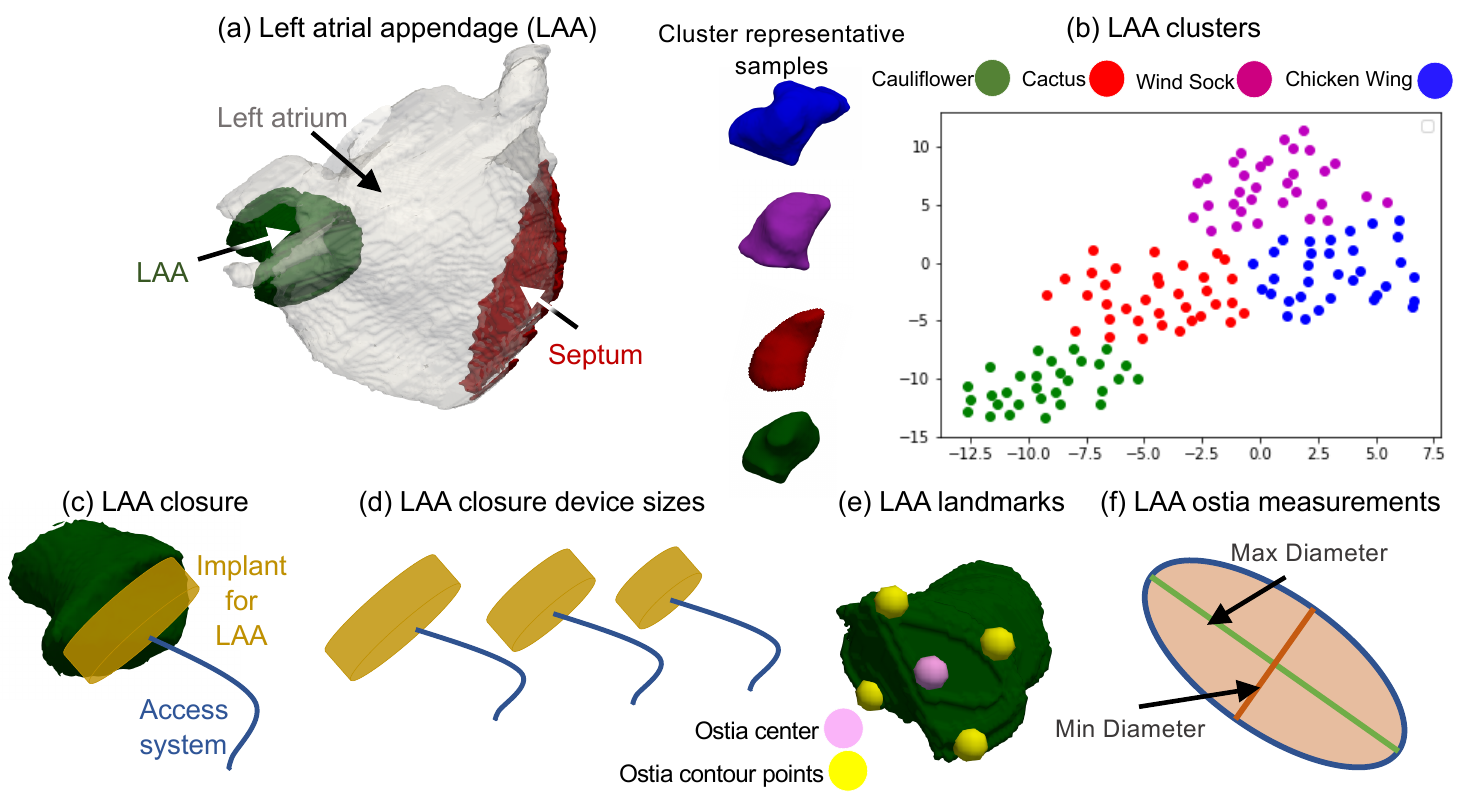}}
    \caption{LAA Anatomy. (a) LAA is a sack-like structure in the human heart; (b) LAA morphology is categorized into four types: chicken wing, wind sock, cactus, and cauliflower \cite{wang2010}. A 2D projection of the clustered LAA shapes from signed distance transform images using t-distributed stochastic neighbor embedding (t-SNE); (c) LAA closure is performed using an implant device through interatrial septum using an access system; (d) LAA closure device sizes available;
    (e) LAA ostia landmarks estimated to measure LAA ostia; and (f) LAA ostia measurements computed from the landmarks by fitting an ellipse. }%
    \label{LAA}
\end{figure}
%-------------------------------------------------------------------------------------
\subsubsection{Implant design and selection -- LAA closure}
%-------------------------------------------------------------------------------------

The left atrial appendage (LAA) is a small sack-like structure in the human heart. In atrial fibrillation (AF) patients, blood clots can form due to irregular heartbeat (i.e., arrhythmia). LAA can be one of the sources for thrombus formation and may be responsible in circulating the blood clots through the body, causing stroke in AF patients \cite{regazzoli2015left}. To reduce the risk of stroke, clinicians occlude the appendage using a closure device (i.e., an implant) (\figurename~\ref{LAA}(a)) \cite{regazzoli2015left}.
LAA morphology is complex and categorized into four types \cite{wang2010}: cauliflower, chicken wing, wind sock, and cactus (\figurename~\ref{LAA}(b)), and hence closure implants are available in various sizes (\figurename~\ref{LAA}(d)) \cite{romero2014left}. 
A clinician typically selects an appropriate device size by examining the patient-specific LAA morphology \cite{wang2010}. Nonetheless, such examination entails significant manual effort for marking relevant anatomical landmarks and measurements, and thereby could lead to subjective and error-prone decisions. Inappropriate device selection would lead to an incomplete LAA closure that is worse than no closure \cite{regazzoli2015left}.
SSM could thus provide an automated approach for developing less subjective categorizations of LAA morphology and anatomical measurements that can be used for more objective clinical decisions regarding suitability for LAA closure. SSM could further assist in designing more accurate, representative implant sizes for different LAA morphologies.

% For a selected implant type, various sizes are supported to fit a patient-specific LAA \cite{romero2014left}. A population-level study of the LAA shapes and LAA ostia using statistical shape modeling can provide objective decisions on the LAA closure implant design and selection process. 

\begin{figure}%[ptb!]
    \centerline{\includegraphics[width =0.95\linewidth]{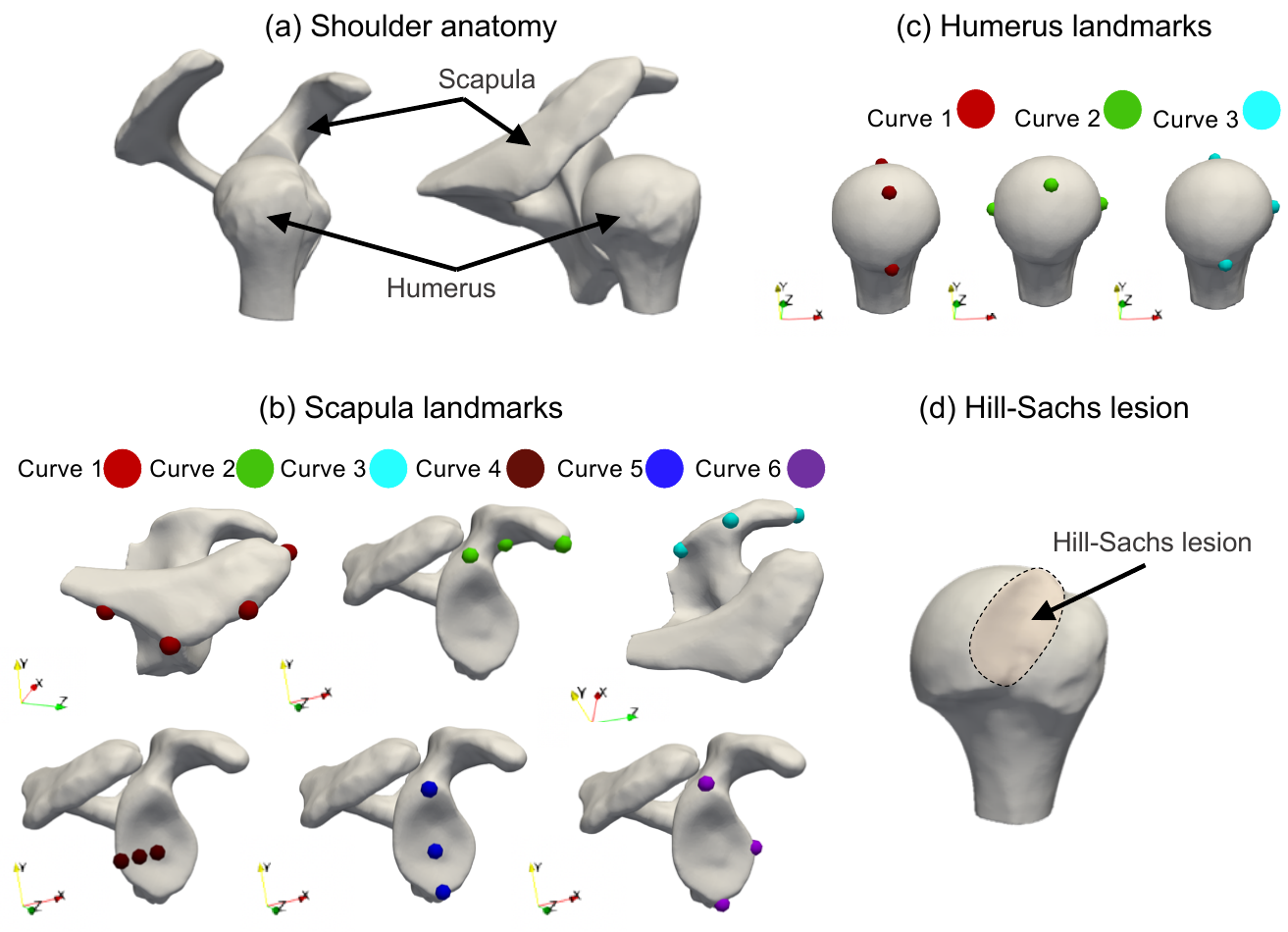}}
    \caption{The human shoulder in SSM applications. (a) The cup-like glenoid of the scapula is the articulating surface for the ball-like humeral head; (b) scapula landmarks obtained for six curves for landmarks inference; (c) humerus landmarks obtained for three curves for landmarks inference; and (d) A Hill-Sachs lesion is formed in the humeral head via compression against the glenoid rim during a shoulder dislocation.}%
    \label{shoulder}
\end{figure}

%-------------------------------------------------------------------------------------
\subsubsection{Surgical planning -- Total shoulder arthroplasty}
%-------------------------------------------------------------------------------------

The scapula is part of the shoulder girdle and has shallow concave glenoid upon which the quasi-spherical humeral head articulates (\figurename~\ref{shoulder}(a)).
The glenohumeral joint can be impaired and worn as seen in osteoarthritis. In these cases, joint replacement with a prosthetic implant, the anatomic total shoulder arthroplasty (aTSA), can reduce pain and restore the normal function of the shoulder joint. In aTSA, restoration of the glenohumeral joint to a nonpathologic state aims to obtain balanced forces on the glenoid and prosthetic components to maintain joint stability and improve the overall shoulder function.
Because of the large anatomic variability of the glenoid \cite{de2010reliability}, no consensus exists on which anatomical references should be used intraoperatively to restore the native glenoid. The inferior section of the glenoid has been found to be the most consistent, and was therefore proposed as a reference. The landmarks defining the native glenoid (\figurename~\ref{shoulder}(b) bottom row) are manually defined on the glenoid and are expert-driven, and thereby their identification can be subjective and error-prone. A patient-specific landmark inference of the scapula can be automated using SSM by relating subject-specific metrics to population-level metrics. Hence, SSM could assist in the restoration of the glenoid plane by providing an objective, automated solution for estimation of landmarks. 

%-------------------------------------------------------------------------------------
\subsubsection{Surgical planning -- Reverse total shoulder arthroplasty}
%-------------------------------------------------------------------------------------

Reverse shoulder arthroplasty is a good treatment option in shoulder pathology with dysfunction of the rotator cuff muscles \cite{saltzman2010method}, including cuff tear arthropathy, irreparable cuff tears, or proximal humerus fractures. In this surgical process, the ball-like structure (i.e., humerus) and socket-like structure (i.e., scapula) are interchanged, hence “reversing” the anatomy of the shoulder. By distalizing and medializing the glenohumeral center of rotation (COR), %(curve 1 landmark at the center of sphere-like humerus head), 
the lever arm of the deltoid muscle is increased so that it can take over shoulder function from the deficient rotator cuff. Lateralization of the humerus without changing the COR can also optimize muscle tension. On the other hand, too much COR lateralization or distalization can lead to bony impingement between the humerus and scapula, nerve lesions, or stress fractures of the scapula.  This interplay amongst range of motion, implant stability, and avoidance of complications is determined by the design of the implant and the clinician’s expertise. SSM could thus automate the inference of optimal COR and landmarks of the humerus to assist in better implant design and implant configuration selection. Furthermore, SSM could also improve the surgical process by objectively characterizing patient-level variability.

\begin{figure}%[ptb!]
    \centerline{\includegraphics[width = 1\linewidth]{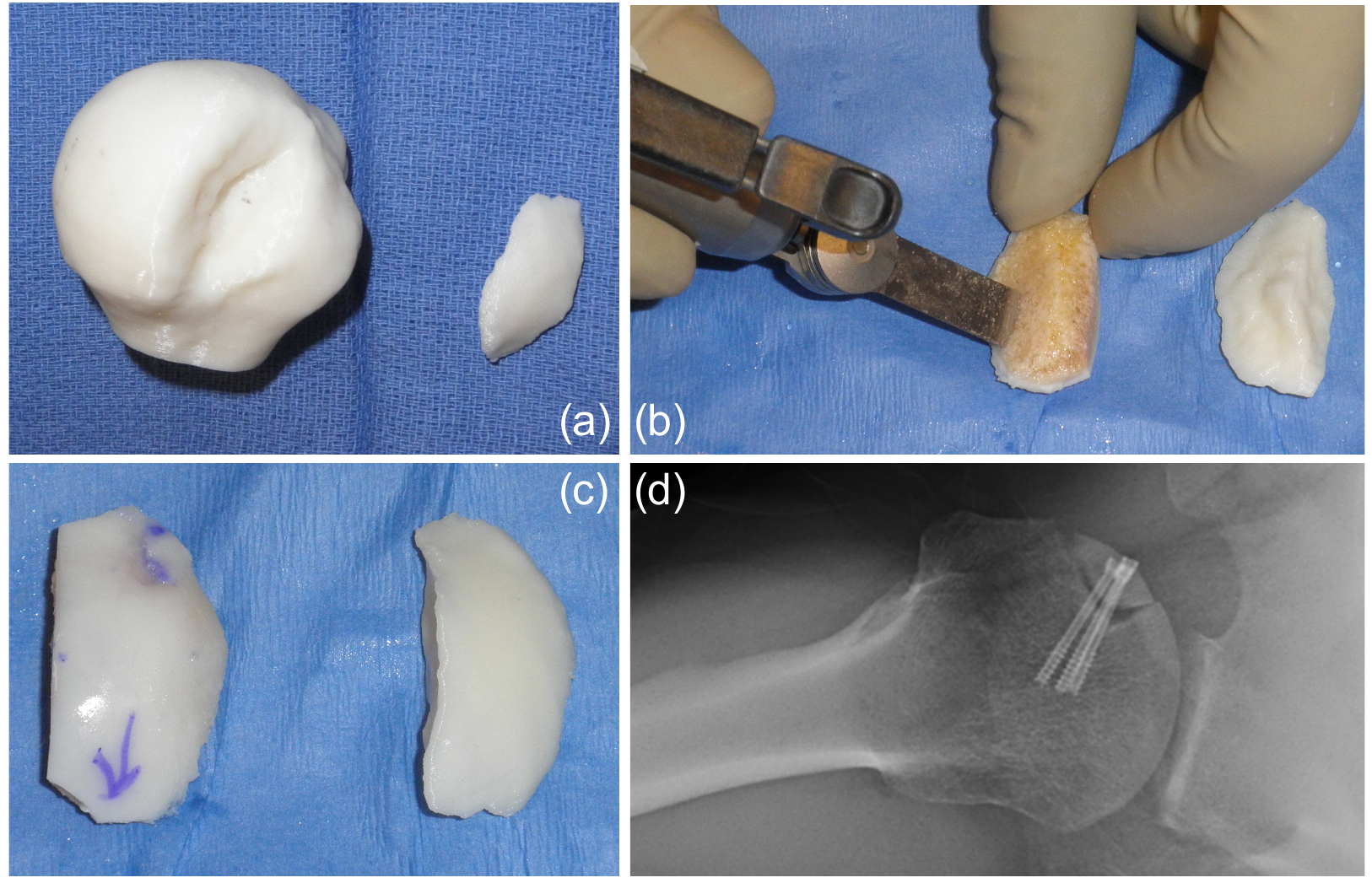}}
    \caption{Hill-Sachs bone grafting. (a) A 3D printed model of a humeral head with a Hill-Sachs defect and a 3D model of the missing bone that fills the void; (b) Shaping a bony allograft to match the size, shape, and orientation of the 3D model; (c) The final graft (left) compared with the 3D template (right); and (d) Postoperative radiograph of the graft in the shoulder. }%
    \label{humerus-illustration}
\end{figure}

%-------------------------------------------------------------------------------------
\subsubsection{Bone grafting -- Hill-Sachs lesion}
%-------------------------------------------------------------------------------------

In cases of shoulder dislocation, the humeral head slips out of the shoulder socket and becomes compressed against the rim of the glenoid, which may lead to compression fractures on the humeral head, also known as a Hill-Sachs lesion (\figurename~\ref{shoulder}(d)). Large Hill-Sachs lesions have a high risk of recurrent shoulder instability, leading to impaired shoulder function and debilitating pain \cite{provencher2012hill}. In cases of large Hills-Sachs lesions, bone grafting of the lesion has been suggested as a viable treatment option. The lesion characteristics are typically evaluated preoperatively on 2D CT-scans. During surgery, measurements are reevaluated using a ruler to choose the fresh frozen allograft that best fits into the defect. Translating this information into a 3D printed model (\figurename~\ref{humerus-illustration}(a)) provides the surgeon with a hands-on template with which to properly template the allograft. Cuts on the allograft are made to shape the graft until it fits the lesion properly (\figurename~\ref{humerus-illustration}(b)). This entire process is performed by trial and error and can vary based on the expertise of the clinician. SSM could assist in the systematic evaluation of the lesion, the lesion depth, and the objective characterization of the filling void to enable objective decisions for sizing and shaping the bone graft.

\begin{figure}%[ptb!]
    \centerline{\includegraphics[width =0.87\linewidth]{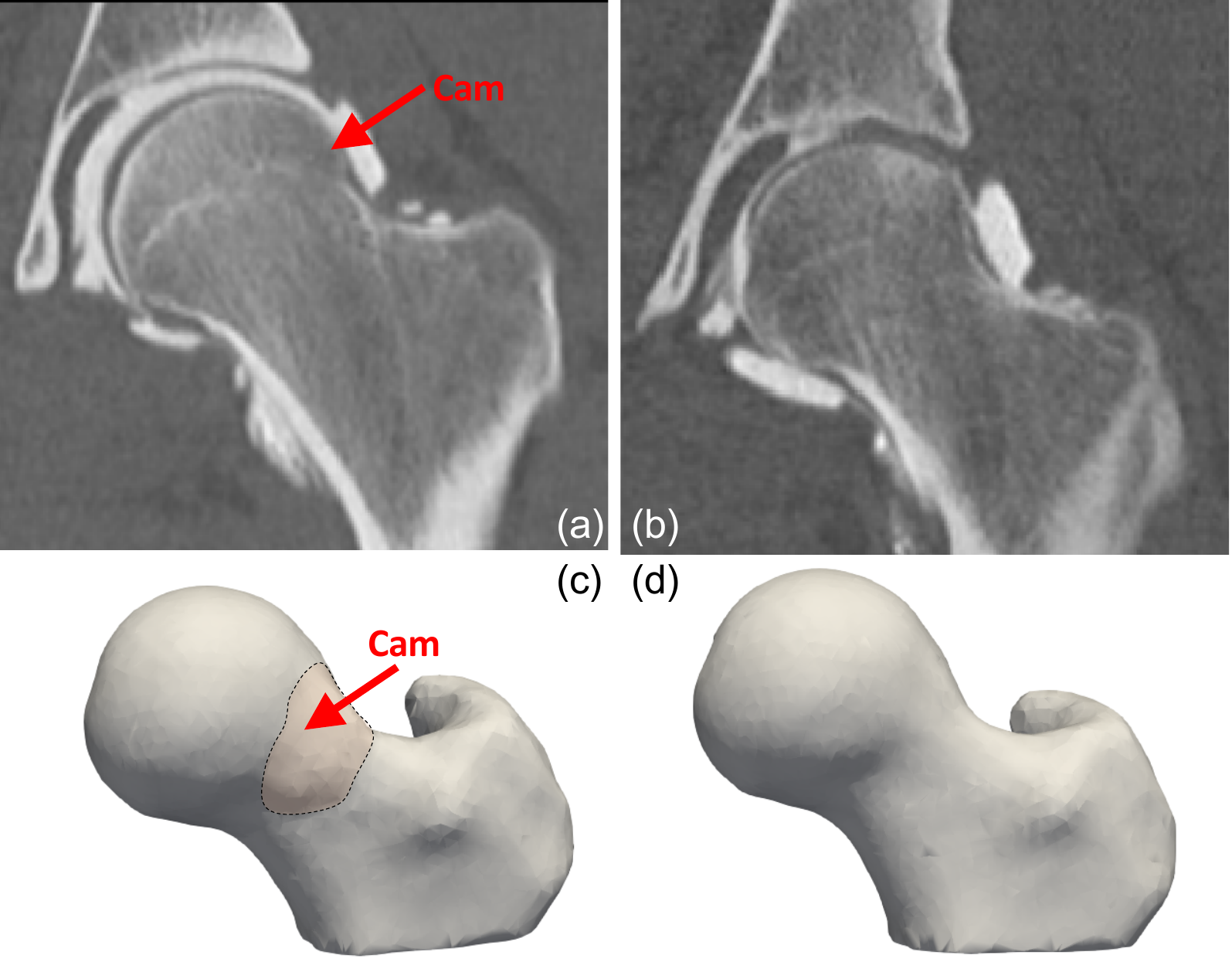}}
    \caption{Cam-type FAI lesion. (a) A CT scan of cam-type FAI femur (an extra bone growth on the femoral head); (b) A CT scan of a control femur ; (c) A 3D segmented and preprocessed femur shape having cam-type FAI; and (d) A 3D segmented and preprocessed control femur. }%
    \label{femur-illustration}
\end{figure}

%-------------------------------------------------------------------------------------
\subsubsection{Bone resection - cam-type FAI lesion}
%-------------------------------------------------------------------------------------

The hip is a ball-socket joint, with the femoral head acting as a ball, and the acetabulum (a component of the pelvis bone) acting as the socket. Femoroacetabular impingement (FAI) occurs when there is extra bone growth along one or both of the bones that form the hip joint (\figurename~\ref{femur-illustration}(a)), which thereby hampers smooth movement. Over time, this abnormal contact can cause damage to the labrum, which is a fibrocartilagenous tissue structure that surrounds the bony rim of the acetabulum. Patients with lesions on the femoral head and head-neck junction are diagnosed with cam-type FAI. Cam is a specific type of FAI in which the bone growth occurs to the femoral neck, i.e., the femoral head does not remain round due to a formed bump, reducing the clearance between the femur and the pelvis (\figurename~\ref{femur-illustration}(c)). Cam-type morphology is believed to cause abnormal motion; notably, rotation of an aspherical femoral head within a relatively spherical socket likely causes the femur to lever-out, in turn leading to high shear stresses on cartilage and the acetabular labrum, leading to tears, fibrillation, and chronic inflammation.
In cam-type FAI patients, the extra bone growth is removed through a surgical resection. Underestimating the resection depth can lead to revision surgery, whereas overestimating the resection depth can lead to hip fractures. Clinicians estimate the cam lesion and the resection depth through inspection of 2D radiographs and visual inspection at the time of surgery.  However, these approaches are only semiquantitative, and may result in over or underestimation of the areal extent and magnitude of the deformity. SSM can automate the detection of the lesion and resection depth, resulting in fewer cases of revision hip arthroscopy.

\begin{figure*}[hbt!]%[p!]
    \centerline{\includegraphics[width=0.95\linewidth]{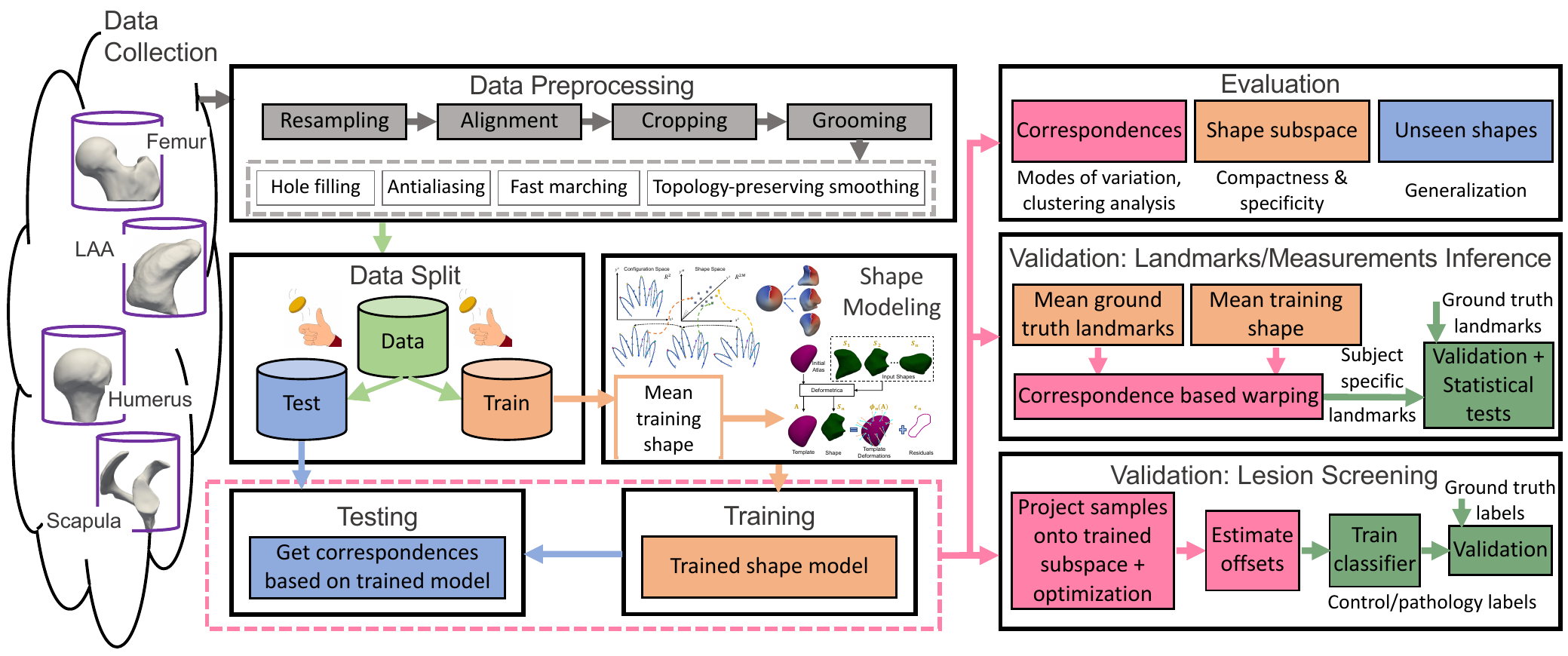}}
    \caption{SSM evaluation and validation frameworks}%
    \label{EvalValPipeline}
\end{figure*}

\begin{figure}[hbt!]%[ptb!]
    \centerline{\includegraphics[width=1\linewidth]{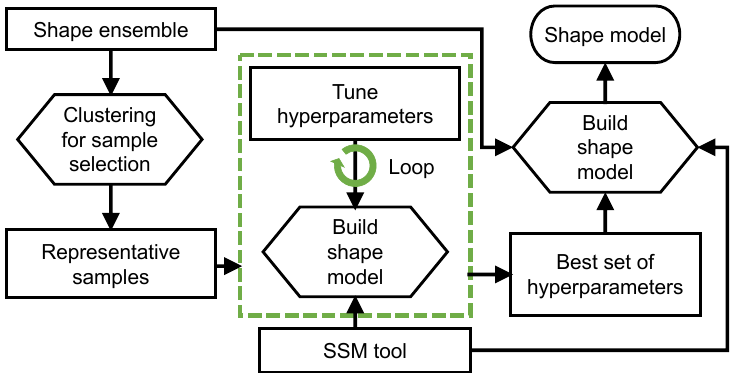}}
    \caption{Hyperparameters tuning. A representative subset is selected using clustering to generate the shape model in an efficient manner.}%
    \label{prep-sample-sel}
\end{figure}

%-------------------------------------------------------------------------------------
\section{Methods: Evaluating and validating SSM tools }\label{eval_val}
%-------------------------------------------------------------------------------------

The assessment of an SSM tool is a multifaceted process where no single metric captures all performance aspects of the resulting shape models. Hence, we present systematic evaluation and validation frameworks (\figurename~\ref{EvalValPipeline}) to assess the point correspondences obtained from different SSM tools. The \textit{evaluation framework} intrinsically assesses the quality of the shape model when the ground-truth correspondences are unavailable. The \textit{validation framework}, on the other hand, is performed in the context of clinical applications where some ground-truth information, extrinsic to the shape model, is available. These frameworks can be applied to any SSM tool, beyond those considered here in this paper.

The \textit{common} steps in the proposed evaluation and validation frameworks are as follows: (1) data collection; (2) data preprocessing; (3) data split; and (4) shape modeling. The \textit{data collection step} entails gathering shape instances (i.e., segmented anatomies as binary images) %(e.g., LAA shape) 
from the population of interest %(e.g., AF partients) 
for statistical analysis. %Shapes are typically given/collected as binary segmentations.

The \textit{data preprocessing step} includes the following: closing small holes in the given segmentations; resampling volumes to have isotropic voxel spacing; antialiasing to remove the staircase effect on the image contours due to discretization \cite{whitaker2000reducing}; aligning center of mass; rigidly aligning shapes using the ensemble mediod as a reference and the advanced normalization tools (ANTs) \cite{avants2014insight} for registration; cropping using the largest bounding box that encapsulates all shape samples to remove the unnecessary background that can slow down the correspondence estimation; fast marching to convert segmentations to signed distance transforms; and topology-preserving smoothing. The preprocessed segmentations are then converted to the appropriate data type needed for each SSM tool (e.g., label maps or surface meshes).

The \textit{data split step} randomly selects samples without replacement to form training and test subsets. The training samples are used to train the shape model, and the testing samples are used for validation. Importance sampling (clustering the data and randomly selecting samples from each cluster) ensures that testing and training subsets are similarly distributed and avoids the bias of the random split in the analysis.

The \textit{shape modeling step} estimates surface correspondences across the training samples using different SSM tools. Each tool (in particular, ShapeWorks and Deformetrica) has a set of algorithmic hyperparameters that need tuning. The hyperparameter tuning is performed on a representative subset of the training samples using K-mediods (\figurename~\ref{prep-sample-sel}). The models resulting from different hyperparameters parameters are compared qualitatively based on two criteria (since ground-truth correspondences are unavailable): (a) correspondence points are evenly spaced to cover the entire geometry; and (b) points are in good correspondence across the training data by inspecting their neighboring correspondences. The best set of hyperparameters is then used for training the shape model on the entire training subset. The trained shape models from SSM tools are used for both evaluation and validation.

%-------------------------------------------------------------------------------------
\subsection{SSM evaluation}\label{sec:ssm-eval}
%-------------------------------------------------------------------------------------

We use quantitative and qualitative metrics to evaluate shape models when ground-truth correspondences are not available. 
% \figurename~\ref{EvalValPipeline} illustrates the proposed evaluation framework.

\begin{figure}%[ptb!]
    \centerline{\includegraphics[width=0.75\linewidth]{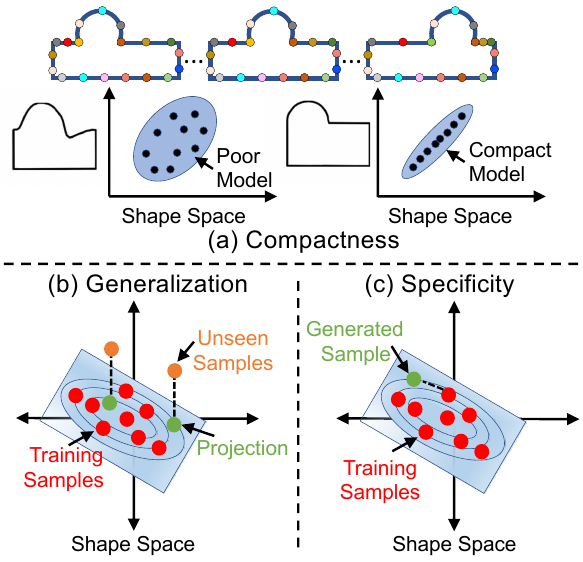}}
    \caption{Quantitative evaluation metrics. (a) A good shape model can encode the shape variability with fewer degrees of freedom; (b) a good shape model can spread between and around the training shapes to represent the unseen shapes; (c) a good shape model can generate plausible shapes.}%
    \label{eval_quant}
\end{figure}

%-------------------------------------------------------------------------------------
\subsubsection{Quantitative evaluation metrics}
%-------------------------------------------------------------------------------------

We adopt the quantitative metrics of compactness, generalization, and specificity \cite{davies20023d} to assess different aspects of a shape model. These measures are functions of the number of modes of variation  $K \in \{1,\ldots, \operatorname{min}(N,dM)\}$ that are computed by PCA on correspondences, where $N$ is the number of training shapes, $d$ is the dimension of the configuration space, and $M$ is the number of correspondences.

%-------------------------------------------------------------------------------------
\vspace{0.05in}\noindent\textbf{Compactness.} Although high-dimensional, the shape space can be parameterized by a low-dimensional subspace (defined by eigenvectors and associated eigenvalues) that explains the shape variability. 
A compact shape model can thus explain such variability with fewer parameters, and the more compact a model is, the better (\figurename~\ref{eval_quant}(a)). 
A compactness measure echoes the Occam's razor principle; ``a simple explanation is more likely to be better than a complicated explanation."
Compactness can be computed as $C(K)= \sum_{j=1}^{K} \lambda _j$ \cite{munsell2008evaluating}, 
where $K$ indicates the number of eigenvectors to explain the shape variability, and $\lambda _j$ indicates the eigenvalue of the $j-$th mode. 
For two shape models $A$ and $B$, if the compactness values are $C_A(K) < C_B(K)$ for one or more values of $K$, then shape model $A$ is said to be more compact than shape model $B$.

%-------------------------------------------------------------------------------------
\vspace{0.05in}\noindent\textbf{Generalization} quantifies whether the probability density function learned by the shape model is able to spread between and around the given training shapes (\figurename~\ref{eval_quant}(b)).
The generalization metric, denoted as $G(K)$ can be computed via a leave-one-out cross-validation as follows \cite{munsell2008evaluating}:    
%Consider $N-$shapes each with $M$ correspondences in a $d-$dim\-e\-n\-s\-i\-onal configuration space. 
Consider a shape vector, $\mathbf{z}_n \in \mathbb{R}^{dM}$, where $n \in \{1,...,N\}$, which is left out from the $N$ shape vectors, and a shape model that is obtained from the rest of the $N-1$ shape vectors. The left-out sample is not considered as part of the SSM correspondence estimation. Generalization can thus be quantified as $G(K)=\frac{1}{N} \sum_{n=1}^{N}\varepsilon_n(K)$, where $\varepsilon _{n}(K)=\lvert\lvert \mathbf{z}_{n}(K) - \mathbf{z}_n \lvert\lvert^2$ is the approximation error using the squared Euclidean distance when using the first $K$ eigenvectors to represent the left-out shape instance. %, repeated and averaged for all $n \in \{1,...,N\}$.
For two shape models $A$ and $B$, if the generalization values are $G_A(K) < G_B(K)$ for one or more values of $K$, then shape model $A$ is said to be more general, representing unseen shapes, than shape model $B$. %Thus, the more general a shape is, the better it can represent unseen shapes. 

%-------------------------------------------------------------------------------------
\vspace{0.05in}\noindent\textbf{Specificity} is the ability of the shape model to generate new, but valid, instances of shapes by constraining the variability in the shape space such that only legal/plausible shapes can be generated (\figurename~\ref{eval_quant}(c)).
Specificity can be quantified by randomly generating $J$ (a large number of) samples $\mathbf{z}(K)$ from the shape space using the first $K$ eigenvectors and eigenvalues, assuming a multivariate normal distribution, and computing the Euclidean distance to the closest training sample $\mathbf{z'}$.
Specificity is computed as $S(K) = \frac{1}{J} \sum_{j=1}^J  \lvert\lvert  \mathbf{z}_j(K) -  \mathbf{z}_{j}^{'} \rvert\rvert ^2$ \cite{munsell2008evaluating}.
% $S(K) = \frac{1}{J} \sum_{j=1}^{J} \lvert\lvert \mathbf{z}_j(K) - {\mathbf{z}_{j}^'}\rvert\rvert^2$ \cite{munsell2008evaluating}.%2 
%
For two shape models $A$ and $B$, if the specificity values are $S_A(K) < S_B(K)$ for one or more values of $K$, then shape model $A$ is said to be more specific, generating more realistic samples, than shape model $B$. %Thus, the more specific a shape model is, the better it can generate more realistic shapes. 

%-------------------------------------------------------------------------------------
\subsubsection{Qualitative evaluation metrics}
%-------------------------------------------------------------------------------------

The qualitative assessment of shape models is performed using modes of variation and cluster analysis. The modes of variation may reflect clinically relevant variations/patterns. For instance, the anterior-posterior dilation of the left atrium shape is found to be statistically correlated with the severity of atrial fibrillation \cite{cates2014computational}. Clustering is an approach to find groups in a population that are as distant as possible while ensuring the samples within a given group to be as similar as possible. Shape populations under analysis in clinical applications may exhibit natural clusters, different levels of illness, and disease progression. For instance, clustering analysis of the left atrium with different pulmonary veins branching might reveal clusters linked to atrial fibrillation pathology \cite{cates2014computational}. A shape model is assessed by the ability to discover such hidden patterns in the shape class of interest.

\begin{figure}%[ptb!]
    \centerline{\includegraphics[width=1\linewidth]{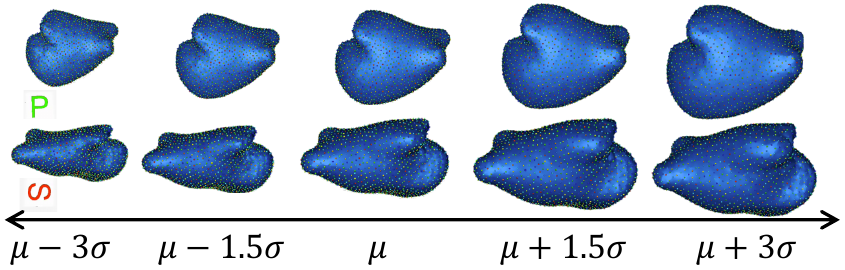}}
    \caption{Left atrium first dominant mode of variation encoding variability in the size of the left atrium in the population (superior and posterior views). Size is not factored out in the left atrium analysis as the left atrium shape (anterior-posterior dilation) is found to be statistically correlated with the severity of atrial fibrillation \cite{cates2014computational}. }%
    \label{ModesofVariation}
\end{figure}

%-------------------------------------------------------------------------------------
\vspace{0.05in}\noindent\textbf{Modes of variation:}
PCA on the point correspondences generated by SSM provides a ranking on the uncorrelated modes of shape variation based on the amount of variance explained (quantified by eigenvalues) relative to the total variance. The modes that explain maximum shape variability are called \textit{dominant} ones. For instance, size is a common dominant mode of variation (\figurename~\ref{ModesofVariation}) in several anatomies, but in few studies, the size variation may be factored out for different purposes (e.g., if not considered as a biological factor). 
In clinical applications, the modes of variation encoded by a shape model can help to objectively characterize normal deformities \cite{harris2013statistical,jacxsens2019coracoacromial},, discover localized pathologies (i.e., abnormalities) in anatomies \cite{atkins2017evaluation,atkins2017quantitative,jacxsens2019coracoacromial}, and identify the severity of a disease \cite{atkins2017quantitative}. Shape models are qualitatively assessed based on the ability to discover clinically relevant modes of variation in the shape class of interest.

%-------------------------------------------------------------------------------------
\vspace{0.05in}\noindent\textbf{Cluster analysis:}
Clustering can discover hidden patterns/groups in the data. In clinical applications, such patterns can assist in morphological classification \cite{goparaju2018evaluation}, disease diagnosis \cite{khanmohammadi2017improved}, and treatment planning \cite{soler2016cluster}. Here, clustering analysis is performed on the point correspondences to assess the ability of a shape model to discover natural clusters.
The inherent number of clusters in a dataset is discovered using the elbow method \cite{hardy1994examination}, which quantifies the percentage of variance explained as a function of the number of clusters found in the data. The first few clusters are expected to explain significant variance, but by adding more clusters, the marginal gain in the explained variance will drop, resulting in an \textit{elbow}. The input shapes and the number of clusters are then provided as input to a clustering algorithm (e.g., K-means, K-medoids) to assign the input shapes to clusters.
For instance, using the elbow method, four clusters, corresponding to the LAA morphological classes, were found in the LAA shape ensemble (\figurename~\ref{LAA}). Here, we used signed distance transform images to serve as a baseline, and the ground-truth cluster labels (i.e., morphology class) were obtained from a clinical expert.
To qualitatively assess a shape model, the point correspondences are clustered to obtain SSM tool-specific clusters. The mean cluster shapes from the ground-truth labeling are then obtained to compare with the clustering results from each shape model. 
This qualitative assessment informs the performance of a shape model in discovering the inherent clusters in the input data.  

%-------------------------------------------------------------------------------------

%-------------------------------------------------------------------------------------
\subsection{SSM validation}
%-------------------------------------------------------------------------------------

We propose two validation frameworks, namely anatomical \textit{landmarks/measurement inference} and \textit{lesion screening}, respectively, where relevant ground-truth (e.g., manually annotated anatomical landmarks) for the validation is obtained from clinical experts. The validation frameworks add two more steps to the steps outlined in Section \ref{eval_val}, validation and statistical tests, which are detailed below for the two proposed frameworks.

%-------------------------------------------------------------------------------------
\subsubsection{Landmarks/measurements inference}\label{methods_landmarks}
%-------------------------------------------------------------------------------------

SSMs can be used to automate the inference of patient-specific anatomical morphometrics such as anatomical landmarks and measurements by defining such morphometrics on the mean shape of a model and using the correspondences to map these morphometrics to the patient space. In this work, patient-specific anatomical landmark estimation is performed for the scapula and the humerus anatomies to assist motion tracking and surgery planning of shoulders. Moreover, estimating patient-specific anatomical measurements is performed for the LAA anatomy to assist in LAA closure device design and selection. The subjective decisions involved in these clinical applications can be reduced by leveraging SSM. 

Given a pretrained shape model, landmark/measurements inference is performed as follows:
Ground-truth landmarks are manually annotated by an expert or with guidance from an expert. The point correspondences for each test sample are then obtained using the shape model learned during the training process. For ShapeWorks, the mean training shape is provided as an initialization for each test sample, where the correspondence optimization is performed only on the test sample. % point correspondences to match the test sample shape. 
For Deformetrica, a deterministic atlas method is used to generate the point correspondences for each test sample by providing the input atlas as the trained output template. For SPHARM-PDM, which follows a pairwise correspondence method, the correspondence generation is the same for train and test samples.

For ShapeWorks and Deformetrica, the patient-specific landmarks are warped from the subject space to the mean space using thin plate splines (TPS) \cite{bookstein1989principal} to compute the mean warped landmarks. 
For SPHARM-PDM, the landmarks on the mean shape are manually annotated as the tool does not provide correspondences in the subject space. %the local coordinate system of the shapes. 
Using correspondences of the mean shape and the patient-specific anatomy as control points, a TPS warp is built to define a mapping between the mean and patient space where the mean landmarks are warped to patient space to obtain patient-specific, \textit{SSM-predicted} landmarks. The landmark predictions from the SPHARM-PDM are aligned to the patient space using a Procrustes fit \cite{gower1975generalized}. 
For the LAA population, which exhibits natural clustering, the ostium is manually annotated using ParaView \cite{ayachit2015paraview} for every cluster mean shape, and the ostium is warped back to the individual samples belonging to the cluster using correspondences as control points for TPS fitting. Finally, the warped ostia shapes are used to compute the LAA ostia measurements (min and max diameters (\figurename~\ref{LAA})), which can be used for the implant design and selection process.

\textit{Validation} entails comparing the SSM-predicted patient-specific landmarks (using Euclidean distance) and measurements (using absolute differences) against the ground-truth ones (\figurename~\ref{EvalValPipeline}). %The landmark differences are computed using Euclidean distance for the landmarks inference task, and the measurement absolute differences are computed for the measurement inference task.
\textit{Statistical tests} identify whether the landmarks/measurements inferred from the SSM tools are statistically equivalent to the ground-truth, 
% Statistical equivalence tests can 
which can assist in drawing conclusions about the relative performance of SSM tools in a clinical application.
Paired sample t-tests \cite{zar1999biostatistical} are used to compare the distance between coordinates of the ground-truth points and the corresponding predicted points in the 3D space. Power analyses are performed, which will indicate all statistical tests we are planning to perform can reach at least 85\% power for the two-sided tests at the 0.05 test level.
%-------------------------------------------------------------------------------------
\subsubsection{Lesion screening}
%-------------------------------------------------------------------------------------

Lesion screening localizes the abnormal changes in a subject-specific anatomy and classifies the subject's anatomy as a control or a pathology based on the extent of the lesion. Applications for lesion screening considered here are the cam-type FAI lesion in femurs and the Hill-Sachs lesion in the humerus. 
In the cam-type FAI lesion, the extra \textit{bone growth} that forms on the edge of the femoral neck is removed through a surgical resection \cite{atkins2017evaluation} (\figurename~\ref{femur-illustration}). Hill-Sachs lesion is a compressive \textit{bone loss} on the humerus head due to dislocation that is filled through a surgical allograft \cite{provencher2012hill} (\figurename~\ref{humerus-illustration}). Accurate lesion extent identification is the key to the success of these surgeries \cite{atkins2017evaluation}. 

\begin{figure}%[b!]
    \centerline{\includegraphics[width=1\linewidth]{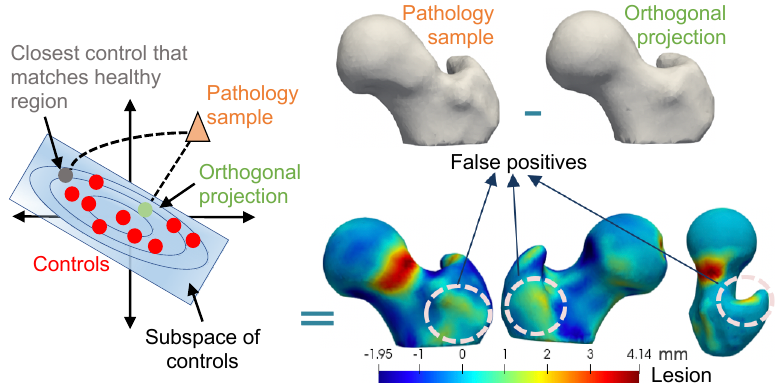}}
    \caption{Pathology sample projection onto controls' subspace. Orthogonal projection of the pathology sample onto controls' subspace fails to determine the accurate representation of the given pathology due to lesion being unsupported by the controls' statistics. Hence, down-weighting the lesion in the projection of the pathology sample via an iterative optimization can help determine the closest control that matches the healthy region.}%
    \label{demoopt}
\end{figure}

SSM can provide an objective characterization of a patient's lesion extent by relating a patient-specific anatomy to the population-level shape statistics of controls. In particular, given a shape model trained on control subjects, a pathologic sample can be represented in the context of the controls population using its closed-form, orthogonal projection onto the PCA subspace of controls. The lesion can then be detected by quantifying the deviation of the pathologic shape from the shape reconstructed based on the model of controls. However, such deviation would result in false positives and fail to determine the accurate representation of the given pathology with respect to the controls' model, primarily because the lesion is a localized abnormal shape change that is not explained by the controls' statistics. If detected or known in advance, the lesion could be discarded, allowing only the healthy parts of the shape to predict the closest control shape to the given pathologic sample, similar to \cite{albrecht2013posterior}. In lesion screening, lesions are not known a priori, and hence representing a pathologic sample with respect to the controls' statistics should down-weight the lesion in the projection of the pathologic shape to reduce false positives in the lesion identification process (\figurename~\ref{demoopt}).

\begin{figure}%[t!]
    \centerline{\includegraphics[width=1\linewidth]{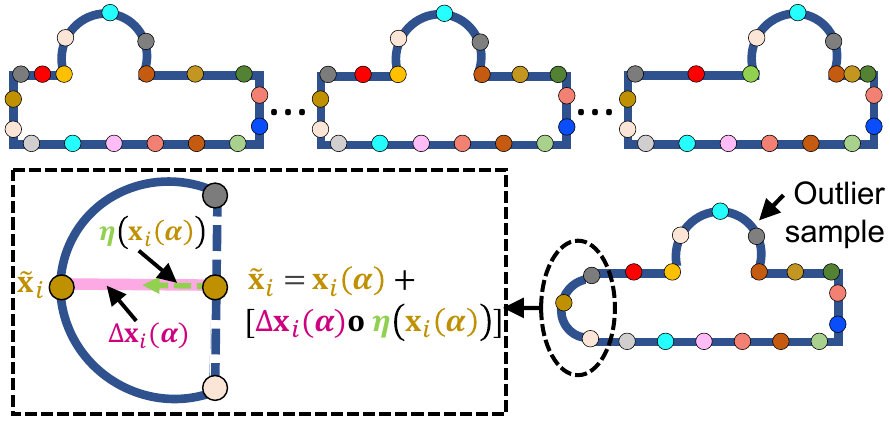}}
    \caption{Illustration of nonorthogonal sample projection optimization using slack variables (surface offsets). Box bump data with an outlier having a bump on the side. The offsets are captured for the points on the side bump alone as the side bump is not present in the rest of the samples.}%
    \label{OffsetsIllustration}
\end{figure}

To reduce false positives, we formulate the projection onto the controls' subspace as an optimization problem that simultaneously estimates the sample's projection and identifies the anatomical regions not supported by the controls shape model. The optimization is formulated using a slack-variables-based approach. In particular, slack variables or surface offsets capture the pointwise differences in the surface normal direction between the pathology sample and the reconstruction of the pathology sample with respect to the controls statistics. 
Since we do not know in advance whether a sample is control or pathology, offsets should be minimal in the case of a control subject, and thereby the solution to this nonorthogonal projection should converge to that of the orthogonal projection. Furthermore, surface offsets should only be nonzero for those point correspondences that belong to the spatial support of the lesion. Hence, the nonorthogonal projection of a pathology sample to the closest control match is formulated as the solution of the following energy function that balances the trade-off between surface reconstruction based on a \textit{pre-trained} shape model and a sparsity inducing regularization for the surface offsets.
\begin{align}\label{eqn:nonortho}
E(\boldsymbol{\alpha}, \Delta \mathbf{x}(\boldsymbol{\alpha})) &=&\sum_{i=1}^{M} \Bigg( {\norm{ \widetilde{\mathbf{x}}_i-\left[\mathbf{x}_i(\boldsymbol{\alpha})+ \Delta \mathbf{x}_i(\boldsymbol{\alpha}) \circ \boldsymbol{\eta} (\mathbf{x}_i(\boldsymbol{\alpha}))\right] }}^2 \nonumber \\
&& ~~~~~~~~~+ \lambda {\norm{ \Delta \mathbf{x}_i (\boldsymbol{\alpha})}}_1 \Bigg)
\end{align}
\noindent where:
\begin{itemize}[noitemsep,topsep=1pt]
\item[$-$] $\widetilde{\mathbf{x}} \in \mathbb{R}^{dM}$ is a pathology sample represented as $M-$correspondence points $\widetilde{\mathbf{x}}_i \in \mathbb{R}^d$ for $i \in \{1,...,M\}$, and $d=3$ for 3D shapes. %$\widetilde{\mathbf{x}_i}$ $\in$ $\mathbb{R} ^{3}$ $i-$th point correspondence. 

\item[$-$] The controls PCA subspace is parameterized by $\Theta = \{\boldsymbol{\mu}, \mathbf{U}\}$, where $\boldsymbol{\mu} \in \mathbb{R}^{dM}$ is the mean shape and $\mathbf{U} \in \mathbb{R}^{dM \times K}$ are the dominant $K-$eigenvectors, i.e., modes of variation, explaining 97\% of the variability in the population. 

\item[$-$] $\boldsymbol{\alpha}$ $\in \mathbb{R} ^{K}$ is the orthogonal projection (i.e., shape parameters) of a pathology sample onto a controls PCA subspace.

\item[$-$] $\mathbf{x}(\boldsymbol{\alpha}) \in \mathbb{R}^{dM}$ denotes the reconstructed pathology correspondences from the controls PCA subspace, computed in closedform as  $\mathbf{x}(\boldsymbol{\alpha}) = \mathbf{U}\boldsymbol{\alpha} +\boldsymbol{\mu}$, with an orthogonal projection of shape parameters $\boldsymbol{\alpha}$. %, where the closed-form orthogonal reconstruction of a pathology from the PCA subspace is computed as $\mathbf{x}(\boldsymbol{\alpha}) = \mathbf{U}\boldsymbol{\alpha} +\boldsymbol{\mu}$.

\item[$-$] $\mathbf{x}_i(\boldsymbol{\alpha})$ $\in$ $\mathbb{R}^{d}$ represents the $i-$th reconstructed correspondence point. To avoid the clutter of notations, we removed the explicit dependency of the reconstructed correspondences on the \textit{pretrained} shape model, i.e., $\mathbf{x}_i(\boldsymbol{\alpha}) = \mathbf{x}_i(\boldsymbol{\alpha |\Theta})$.

\item[$-$] $\boldsymbol{\eta} (\mathbf{x}(\boldsymbol{\alpha})) \in \mathbb{R}^{dM}$ is the surface normal vectors for the correspondences on the pathology reconstruction and $\boldsymbol{\eta} (\mathbf{x}_i(\boldsymbol{\alpha}))$ $\in$ $\mathbb{R} ^{d}$ is the normal vector of the $i-$th correspondence $\mathbf{x}_i(\boldsymbol{\alpha})$. 

\item[$-$] $\Delta \mathbf{x}(\boldsymbol{\alpha}) \in \mathbb{R}^{dM}$ is vector of surface offsets for the correspondence points on the pathology reconstruction. $\Delta \mathbf{x}_i(\boldsymbol{\alpha}) =  \Delta x_i(\boldsymbol{\alpha}) \mathbf{1}_d \in \mathbb{R}^d $ represents a vector of offsets with equal elements $\Delta x_i(\boldsymbol{\alpha})$ for the $i-$th correspondence in the direction of surface normal $\boldsymbol{\eta} (\mathbf{x}_i(\boldsymbol{\alpha}))$ on the shape $\mathbf{x}(\boldsymbol{\alpha})$. 

\item[$-$] $\circ$ denotes elementwise (i.e., Hadamard) product.

\item[$-$] $\lambda$ is the regularization parameter of the sparsity prior on the surface offsets to force zero offsets for regions/samples that are explained by the controls statistics. 

\end{itemize}

\begin{algorithm}
\caption{Nonorthogonal sample projection using slack-variables-based optimization}\label{alg:euclid}
\begin{algorithmic}[1]
\State \textbf{Input:}
(a) Shape sample ($\widetilde{\mathbf{x}}$ with $M-$correspondences), (b) a signed distance transform (SDT) representation for the given sample to compute surface normal vectors $\boldsymbol{\eta} (\mathbf{x}(\boldsymbol{\alpha}))$ for each correspondence in $\mathbf{x}(\boldsymbol{\alpha})$, and (c) controls PCA subspace (mean $\boldsymbol{\mu}$, eigenvectors $\mathbf{U}$ defined by $K$ modes).
\State \textbf{Output:} $\boldsymbol{\alpha}$: sample projection onto controls subspace, and $\Delta \mathbf{x}(\boldsymbol{\alpha})$: pointwise surface offsets.  
\State \textbf{Initialize parameters: } Initial sample projection using PCA orthogonal projection, i.e., $\boldsymbol{\alpha}^{(0)} = \mathbf{U}^T(\widetilde{\mathbf{x}} - \boldsymbol{\mu})$, and set the offset values for each point as $1e-06$. 
\State \textbf{Compute derivatives for $\boldsymbol{\alpha}$: } Compute $\frac{\partial E}{\partial \boldsymbol{\alpha}}$ using (\ref{eqn:gradalpha}).
\State \textbf{Update $\boldsymbol{\alpha}$:} $\boldsymbol{\alpha} ^{(t+1)}=\boldsymbol{\alpha} ^{(t)} - \omega \frac{\partial E}{\partial \boldsymbol{\alpha}}$ if the update reduces the energy function, where $\omega$ is an adaptive learning rate. 
\State \textbf{Reconstruct $\mathbf{x}$: } Compute $\mathbf{x}^{(t+1)}(\boldsymbol{\alpha})=\mathbf{U}\boldsymbol{\alpha} ^{(t+1)}+\boldsymbol{\mu}$.
\State \textbf{Compute surface normals:} Use the gradient of the SDT to compute the surface normals at the updated correspondence points, i.e., $\boldsymbol{\eta} ^{(t+1)} (\mathbf{x}(\boldsymbol{\alpha})) =\boldsymbol{\eta} (\mathbf{x} ^{(t+1)}(\boldsymbol{\alpha}))$.
\State \textbf{Compute derivatives for $\Delta \mathbf{x}(\boldsymbol{\alpha})$: } Compute $\frac{\partial E}{\partial \Delta \mathbf{x}_i(\boldsymbol{\alpha}) }$ for $i=\{1,...,M\}$ using (\ref{eqn:grad-offset}).
\State \textbf{Update $\Delta \mathbf{x}(\boldsymbol{\alpha})$:} $\Delta \mathbf{x}_i ^{(t+1)} (\boldsymbol{\alpha})=\Delta \mathbf{x}_i ^{(t)}(\boldsymbol{\alpha}) - \boldsymbol{\gamma}_i \frac{\partial E}{\partial \Delta \mathbf{x}_i(\boldsymbol{\alpha}) }$ if the update reduces the energy function. Here, $\boldsymbol{\gamma}_i$ is an adaptive learning rate for the $i-$th correspondence point. 
\State \textbf{Repeat steps 4-9 until the maximum number of iterations or convergence are computed as $\frac {\lvert \boldsymbol{\alpha} ^{(t)} - \boldsymbol{\alpha} ^{(t-1)} \rvert}{\lvert \boldsymbol{\alpha} ^{(t-1)} \rvert}$
and
$\frac {\lvert \Delta \mathbf{x}(\boldsymbol{\alpha}) ^{(t)} - \Delta \mathbf{x}(\boldsymbol{\alpha}) ^{(t-1)} \rvert}{\lvert \Delta \mathbf{x}(\boldsymbol{\alpha}) ^{(t-1)} \rvert} <  1e-06 $}

\end{algorithmic}
\end{algorithm}

The energy function in (\ref{eqn:nonortho}) is  minimized using gradient-descent optimization with an  alternating coordinate descent on the parameters $\boldsymbol{\alpha}$ and $\Delta \mathbf{x}(\boldsymbol{\alpha})$. The L2 norm on the difference between the pathology sample and the reconstructed sample is minimized by encoding the differences attributed to the lesion variations not supported by the shape model in the surface/point offsets. The L1 regularization is used to induce sparsity on the offsets by allowing the differences to be captured only for the points not supported by the controls PCA subspace (\figurename~\ref{OffsetsIllustration}). The partial derivatives with respect to $\boldsymbol{\alpha}$ are as follows:
\begin {eqnarray}\label{eqn:gradalpha}
\frac{\partial E}{\partial \boldsymbol{\alpha}} &=& 2 {\left( \sum_{i=1}^{M}\Big\{  \widetilde{\mathbf{x}}_i-[\mathbf{x}_i(\boldsymbol{\alpha})+ \Delta \mathbf{x}_i(\boldsymbol{\alpha}) \circ \boldsymbol{\eta} (\mathbf{x}_i(\boldsymbol{\alpha}))] \Big\} \right )}^T \nonumber \\
&&~~~~~~~\times\left ( \frac{\partial E}{\partial \boldsymbol{\alpha}} \Big\{ - \left [\mathbf{x}_i(\boldsymbol{\alpha})+\Delta \mathbf{x}_i(\boldsymbol{\alpha}) \circ \boldsymbol{\eta} (\mathbf{x}_i(\boldsymbol{\alpha})) \right ]\Big\} \right)
\end{eqnarray}
\noindent Using the closed-form orthogonal reconstruction of a pathology from the PCA subspace, the vector representation of the gradient computation is given as
\begin {equation}\label{eqn:grad-xalpha}
\frac{\partial E}{\partial \boldsymbol{\alpha}} \mathbf{x}(\boldsymbol{\alpha}) = \frac{\partial E}{\partial \boldsymbol{\alpha}} \{ \mathbf{U}\boldsymbol{\alpha} + \boldsymbol{\mu} \} = \mathbf{U},
\end{equation}
where $\frac{\partial E}{\partial \boldsymbol{\alpha}} \mathbf{x}_{i}(\boldsymbol{\alpha}) \in  \mathbb{R} ^{d \times K}$.
In an alternating coordinate descent, surface offsets $\Delta \mathbf{x}(\boldsymbol{\alpha})$ are assumed to be fixed (i.e., lagging) with respect to $\alpha$. The derivative computations of surface normals $\boldsymbol{\eta}$ with respect to $\boldsymbol{\alpha}$ is approximated using finite differences across iteration $(t)$ and $(t-1)$, with a vector representation written as
\begin {equation}\label{eqn:grad-normals}
\frac{\partial E}{\partial \boldsymbol{\alpha}} \left[\Delta \mathbf{x}(\boldsymbol{\alpha}) \circ  \boldsymbol{\eta} (\mathbf{x}(\boldsymbol{\alpha}))\right] = \Delta \mathbf{x} (\boldsymbol{\alpha}) \circ \left (  \frac{\boldsymbol{\eta} ^{(t)} (\mathbf{x}(\boldsymbol{\alpha})) - \boldsymbol{\eta} ^{(t-1)} (\mathbf{x}(\boldsymbol{\alpha}))}{\boldsymbol{\alpha} ^{(t)} - \boldsymbol{\alpha} ^{(t-1)}} \right).
\end{equation}
Gradients from (\ref{eqn:grad-normals}) result in a matrix $\mathbb{R}^{dM \times K}$, which are summed with the gradients from (\ref{eqn:grad-xalpha}) and multiplied with the vectorized form of $[\widetilde{\mathbf{x}}_i-(\mathbf{x}_i(\boldsymbol{\alpha})+ \Delta \mathbf{x}_i(\boldsymbol{\alpha}) \circ \boldsymbol{\eta} (\mathbf{x}_i(\boldsymbol{\alpha}))]$, which is $[\widetilde{\mathbf{x}}-(\mathbf{x}(\boldsymbol{\alpha})+ \Delta \mathbf{x}(\boldsymbol{\alpha}) \circ  \boldsymbol{\eta} (\mathbf{x}(\boldsymbol{\alpha})] \in \mathbb{R} ^{dM \times 1}$, resulting in a $\mathbb{R}^K$ gradient.

For a given correspondence point offset, the partial derivatives of $ \Delta \mathbf{x}_i(\boldsymbol{\alpha})$ are computed as follows:
\begin {eqnarray}\label{eqn:grad-offset}
\frac{\partial E}{\partial \Delta \mathbf{x}_i(\boldsymbol{\alpha})} &=& 2{\Big( \widetilde{\mathbf{x}}_i-[\mathbf{x}_i(\boldsymbol{\alpha})+ \Delta \mathbf{x}_i(\boldsymbol{\alpha}) \circ \boldsymbol{\eta} (\mathbf{x}_i(\boldsymbol{\alpha}))] \Big)}^T \nonumber \\
&& ~~~\times \left( \frac{\partial E}{\partial \Delta \mathbf{x}_i(\boldsymbol{\alpha})} \left \{-\Delta \mathbf{x}_i(\boldsymbol{\alpha}) \circ \boldsymbol{\eta} (\mathbf{x}_i(\boldsymbol{\alpha})) \right \} \right ) \nonumber \\
&& ~~~~ + \frac{\partial E}{\partial \Delta \mathbf{x}_i(\boldsymbol{\alpha})} \lambda {\norm{ \Delta \mathbf{x}_i (\boldsymbol{\alpha}) }}_1.
\end{eqnarray}

\begin{figure}[b!]
    \centerline{\includegraphics[width=1\linewidth]{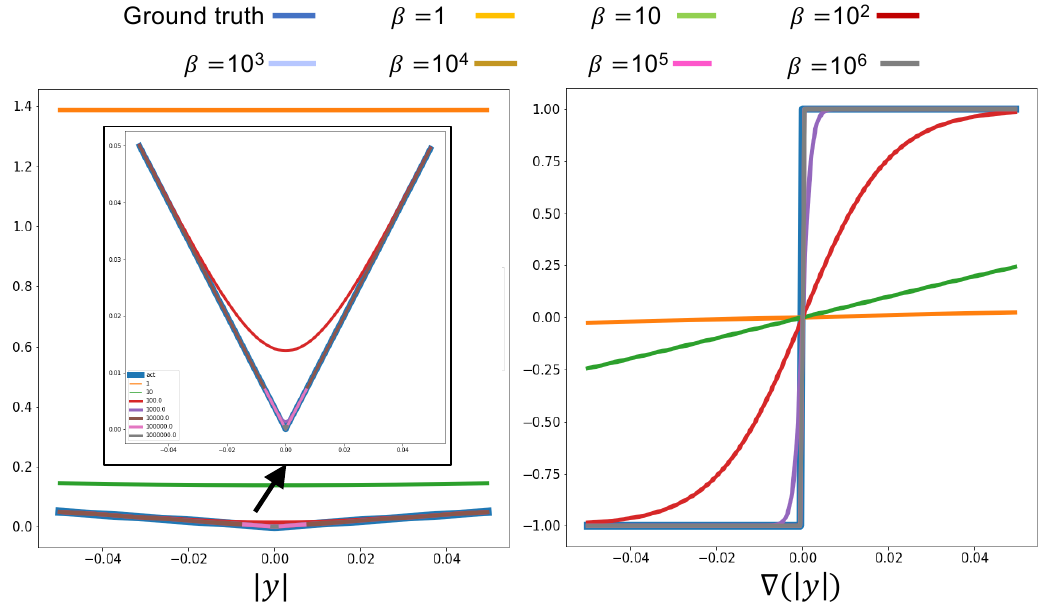}}
    \caption{L1 norm approximation with different $\beta$ values \cite{schmidt2007fast}. $\beta=10 ^6$ results in the approximation that is within a small tolerance of the results produced by constrained optimization methods.}%
    \label{L1normapprox}
\end{figure}

% \begin {equation}
% \begin{matrix}
% \frac{\partial E}{\partial \Delta \mathbf{x}_i(\boldsymbol{\alpha})}= {\Big( \widetilde{\mathbf{x}_i}-(\mathbf{x}_i(\boldsymbol{\alpha})+ \Delta \mathbf{x}_i(\boldsymbol{\alpha}) . \boldsymbol{\eta} (\mathbf{x}_i(\boldsymbol{\alpha}))) \Big)}^T \\. \left( \frac{\partial E}{\partial \Delta \mathbf{x}_i(\boldsymbol{\alpha})} \left \{-(\Delta \mathbf{x}_i(\boldsymbol{\alpha}) . \boldsymbol{\eta} (\mathbf{x}_i(\boldsymbol{\alpha}))) \right \} \right )\\
%                                             + \frac{\partial E}{\partial \Delta \mathbf{x}_i(\boldsymbol{\alpha})} \lambda {\norm{ \Delta \mathbf{x}_i (\boldsymbol{\alpha}) }}_1.

% \end{matrix}
% \end{equation}

\noindent where
\begin {equation}\label{eqn:grad-offset-part1}
\frac{\partial E}{\partial \Delta \mathbf{x}_i(\boldsymbol{\alpha})} \left \{\Delta \mathbf{x}_i(\boldsymbol{\alpha}) \circ \boldsymbol{\eta} (\mathbf{x}_i(\boldsymbol{\alpha})) \right\} = \boldsymbol{\eta} (\mathbf{x}_i(\boldsymbol{\alpha})).
\end{equation}
L1 norm is a non-differentiable penalty. Here, we use a smooth approximation to the L1 penalty consisting of the sum of the integral of two sigmoid functions defined by Schmidt \cite{schmidt2007fast}, where $\beta=10^6$ results in the approximation that is within a small-enough tolerance of the results produced by constrained optimization methods (\figurename~\ref{L1normapprox}).
\begin{equation}\label{eqn:l1normapprox}
\lvert y \rvert \approx \frac{1}{\beta} \left \{ \log \left (1+\exp(-\beta y) \right) + \log \left ( 1+\exp (\beta y)\right ) \right \}.
\end{equation}
% \begin{equation}
% \triangledown(\lvert y \rvert) \approx \frac{1}{(1+\exp(-\beta y)} - \frac{1}{(1+\exp(\beta y))}.
% \end{equation}
Hence, the gradient of the L1 norm approximation can be written as
\begin{eqnarray}\label{eqn:grad-l1norm}
\frac{\partial E}{\partial \Delta \mathbf{x}_i(\boldsymbol{\alpha})} \lambda {\lvert \lvert \Delta \mathbf{x}_i(\boldsymbol{\alpha}) \rvert\rvert}_1 &\approx& \lambda \left(\frac{1}{1+\exp(-\beta \Delta \mathbf{x}_i(\boldsymbol{\alpha}))} \right. \\ \nonumber 
&& ~~~\left.- \frac{1}{1+\exp(\beta \Delta \mathbf{x}_i(\boldsymbol{\alpha}))}\right).
\end{eqnarray}
Equations (\ref{eqn:grad-offset-part1}) and (\ref{eqn:grad-l1norm}) are for a given point correspondence. Considering all the $M$ points on a pathology reconstruction, (\ref{eqn:grad-offset-part1}) results in gradients in $ \mathbb{R} ^ {M \times d}$ when converted from flattened vector in $ \mathbb{R} ^ {dM}$ to 3D points. Equation \ref{eqn:grad-l1norm} obtains gradients in $\mathbb{R} ^ {M }$.  The gradient from (\ref{eqn:grad-offset-part1}) is multiplied by the 
% $(\widetilde{\mathbf{x}_i}-(\mathbf{x}_i(\boldsymbol{\alpha})+ \Delta \mathbf{x}_i(\boldsymbol{\alpha}) . \boldsymbol{\eta} (\mathbf{x}_i(\boldsymbol{\alpha}))))$ that is 
$\widetilde{\mathbf{x}}-(\mathbf{x}(\boldsymbol{\alpha})+ \Delta \mathbf{x}(\boldsymbol{\alpha}) \times \boldsymbol{\eta} (\mathbf{x}(\boldsymbol{\alpha})))$ converted to 3D points in $ \mathbb{R} ^{M \times d}$ and summed up across the dimensions resulting gradients in $ \mathbb{R} ^{M}$. These results are summed with the gradients from (\ref{eqn:grad-offset-part1}) to get the final gradients in $\mathbb{R} ^{M}$.
The gradients obtained above are used to minimize the objective function in an iterative manner using an adaptive learning rate (see Algorithm 1). The parameters that minimize the energy function in (\ref{eqn:nonortho}) are used to compute the closest control to the given pathology sample $\mathbf{x}(\boldsymbol{\alpha})$, and the offsets $\Delta \mathbf{x} (\boldsymbol{\alpha})$ indicate the extent of the lesion.

\begin{figure*}%[ptb!]
    \centerline{\includegraphics[width = 0.96\linewidth]{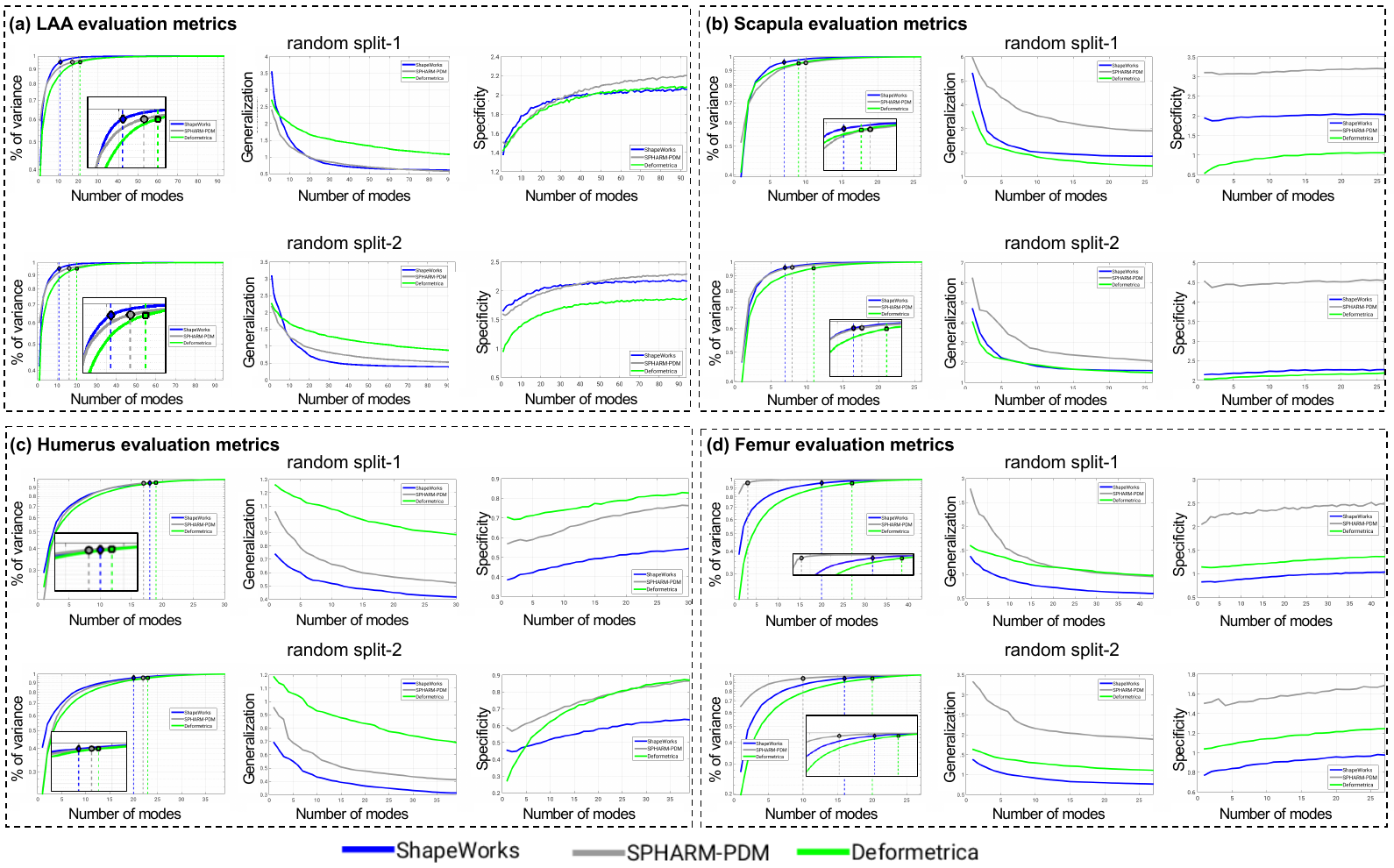}}
    \caption{Compactness (higher is better), generalization (lower is better), and specificity
(lower is better) computed for shape models of (a) LAA, (b) scapula, (c) humerus, and (d) femur random splits.}%
    \label{evaluation_metrics_all}
\end{figure*}

The offsets are used to validate the performance of the SSM tools in identifying the lesion. A qualitative assessment of shape models from SSM tools is performed as follows: (1) the group differences between the original controls, held-out samples, and the reconstructed controls with offsets are not expected to provide any significant differences because the shape model should explain controls variability; and (2) the group differences between the pathology and the reconstructed pathology samples with offsets are expected to inform differences localized to a particular region. The group differences are visualized, and the offset values are assessed across shape models. 

The estimated offsets are also used in a pathology classification task (\figurename~\ref{EvalValPipeline}). A random train and test split is performed on the controls and pathology samples. The offsets of the training samples are fed to a multilayer perceptron classifier, with labels 0 and 1 indicating control and pathology, respectively. The accuracy of the classifier is then obtained by testing the model on the test, held-out samples, consisting of controls and pathology subjects. Multiple train-test splits are performed, and the average accuracy of the classifier is reported for each SSM tool.

%-------
% \input{background_tools}
% \input{background_clinical_applications}
% \input{methods_framework}
% \input{methods_evaluation_quantitative}
% \input{methods_evaluation_qualitative}
% \input{methods_validation_common_steps}
% \input{methods_validation_landmarks}
% \input{methods_validation_lesion}
%-------

\begin{figure*}%[ptb!]
    \centerline{\includegraphics[width = 0.83\linewidth]{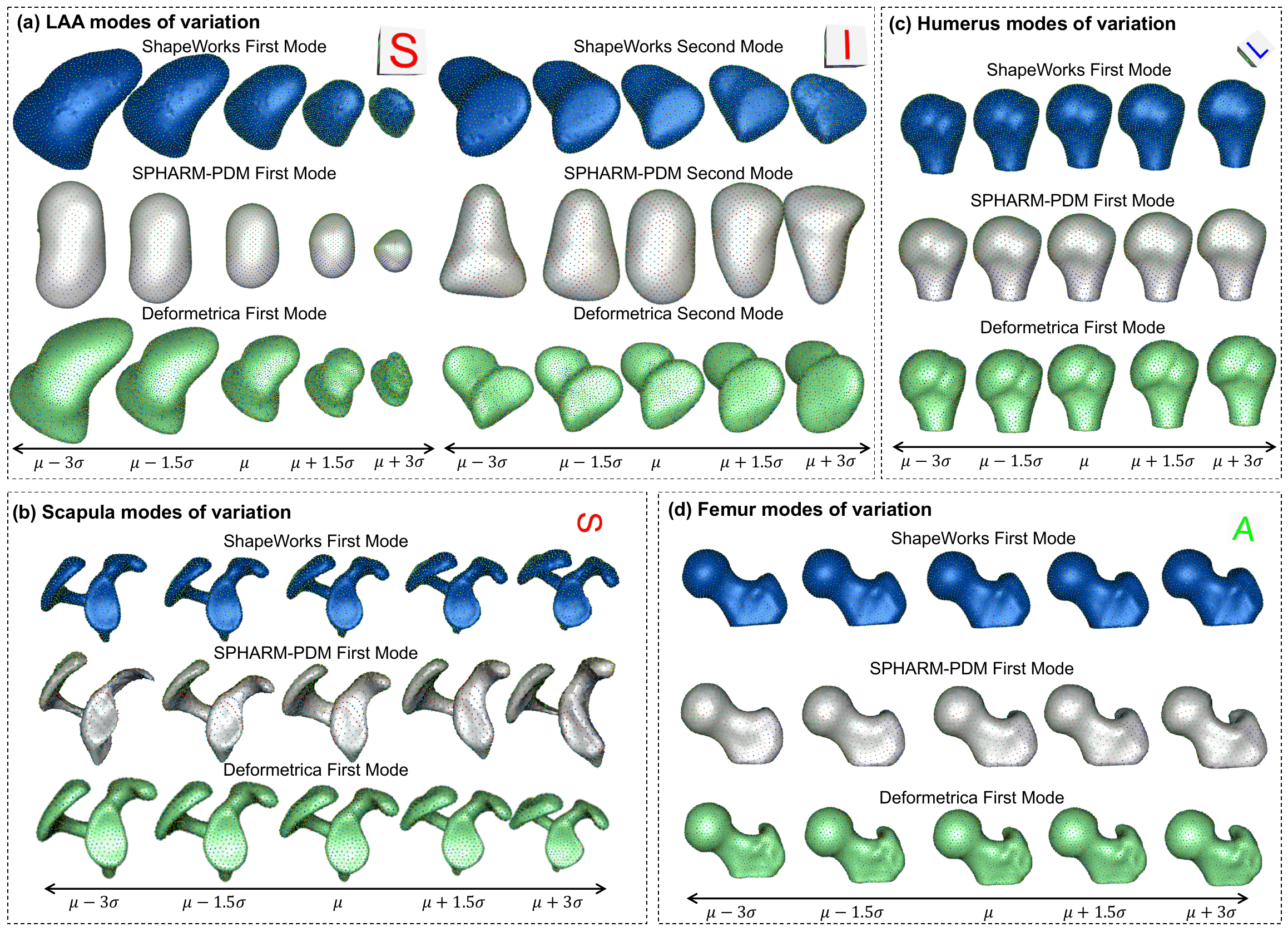}}
    \caption{Shape modes of variation for (a) LAA, (b) scapula, (c) humerus, and (d) femur datasets. (a) Superior (S) and inferior (I) views are shown for LAA. (b) Superior (S) view is shown scapula. (c) Left (L) view is shown for humerus. (d) Anterior (A) view is show for femur.}%
    \label{modes_of_variation_all}
\end{figure*}

\section{Results}

This section presents the evaluation and validation results of the considered SSM tools (Section \ref{ssm-tools}) for a representative set of clinical applications (Section \ref{clinical-apps}) that demonstrate common and important clinical utilities of shape modeling.

%-------------------------------------------------------------------------------------
\subsection{Experimental setup}
%-------------------------------------------------------------------------------------

Here, we cover datasets and training/testing splits for building shape models considered for the benchmark study.

%-------------------------------------------------------------------------------------
\subsubsection{Datasets}
%-------------------------------------------------------------------------------------

\noindent\textbf{Left atrial appendage (LAA).} 
The population study was conducted on 130 LAA images that were retrospectively obtained from the AFib database at the University of Utah. The MRI images were served with a single-handed segmentation by an expert. 
The ground-truth landmarks consisting of five points on the LAA ostium were manually annotated for each LAA sample using Corview (Marrek inc., Salt Lake City, UT), as shown in \figurename~\ref{LAA}(f), and reviewed by a clinical expert.
The segmented binary volumes of LAA were preprocessed with a pipeline involving isotropic resampling (0.625 mm for voxel spacing), antialiasing, center of mass alignment, and rigid alignment. 
The reference image for the rigid alignment was selected as a representative shape of the entire cohort using K-medoids clustering, assuming the entire dataset belongs to one single cluster. 
The clustering process was performed on signed distance transform images. The images were then cropped by estimating the largest bounding box of all the shapes to enable faster processing.

\noindent\textbf{Scapula.}
CT scans and corresponding scapula segmentations of 31 cadaveric control scapulae and 54 scapulae of patients with shoulder instability were obtained from the coracoacromial morphology study in \cite{jacxsens2019coracoacromial}. 
The anatomical landmarks obtained for the scapulae participants under the coracoacromial morphology study were used here for the validation of the landmarks inference. 
The ground-truth landmarks were manually annotated for six curves, as shown in \figurename~\ref{shoulder}(b). A best-fit circle of the glenoid was used for the glenoid landmark annotation. 
The significance of such landmarks is as follows:
Curve 1 landmarks represent the anatomy of acromion, 
curves 2 and 3 landmarks capture the coracoid process, 
curve 4 and 5 landmarks obtain the curvature of the concave articular surface of the glenoid, and
curve 6 landmarks encode the anterior rim of the glenoid to address potential anterior defects.
These landmarks are of interest to address both the glenoid and the coracoacromial anatomy to understand the pathoanatomy and pathomechanics of shoulder instability. 
The data as part of the coracoacromial morphology study \cite{jacxsens2019coracoacromial} were preprocessed as follows-  The left scapulae shapes were mirrored to right scapulae shapes to ensure a consistent orientation of all the shapes in the cohort. Scapulae shapes were aligned to the glenoid-based coordinate system. % and converted to binary segmentations using a spatial partitioning algorithm \cite{atkins2017quantitative}. 
Additional preprocessing steps such as resampling (0.5 mm voxel spacing), antialiasing, and cropping using the largest bounding box were performed for the scapulae shapes, similar to the LAA shapes.

\noindent\textbf{Humerus.}
CT scans and humerus segmentations of 31 cadaveric control humeri and 54 humeri of patients with shoulder instability and a Hill-Sachs lesion were obtained as part of the study in \cite{jacxsens2019coracoacromial}.
The ground-truth landmarks were obtained for three anatomical curves, as shown in \figurename~\ref{shoulder}(c), which encode the morphological information of the humeral head. Information on the articular surface is encoded in curves 1 to 3 (\figurename~\ref{shoulder}(c)). The inference of these landmarks can help in surgical planning.
The data preprocessing steps for the humerus were the same as those for the scapulae.

\noindent\textbf{Femur.}
The femurs data were collected through CT scans of 59 control and 37 FAI patients with cam-type lesions. These scans were obtained as part of the cortical bone thickness study \cite{atkins2017quantitative}. The data preprocessing steps for the femurs were the same as those for the scapulae.

% The CT scans were upsampled to an isotropic voxel size of 0.33 mm to enhance the resolution. Surface meshes for femurs were obtained using the Amira software \cite{stalling2005amira} using a processing pipeline detailed in \cite{atkins2017quantitative}. The meshes were then converted to signed distance transforms and binary volumes.

%-------------------------------------------------------------------------------------
\subsubsection{Train/test splits and shape modeling}
%-------------------------------------------------------------------------------------

\noindent\textbf{LAA.} Using the elbow method \cite{hardy1994examination}, the number of clusters in the LAA dataset was identified as four, matching the LAA morphology classification reported in the literature \cite{wang2010}.
Seventy percent of the samples were selected using random sampling without replacement from each cluster to serve as training data. 
The remaining samples from each cluster were considered as testing data. Two such random train and test splits were sampled to perform the analysis.
For each random split, the training data were fed to each SSM tool to build the shape model. Since ShapeWorks and Deformetrica rely on a groupwise optimization approach, the point correspondences for each of the test samples were obtained by using the mean shape for initialization and fixing the correspondence of the training samples (i.e., the shape model). 
The process of obtaining the correspondences for each test sample from SPHARM-PDM is the same as training due to its pairwise approach.

\noindent\textbf{Scapula.}
Controls and pathology cohorts were used to generate two random splits. 
Split-1 had controls as training data and pathology samples as testing, whereas split-2 was constructed with pathology samples as training data and controls as testing data.
The purpose of these splits is to assess the performance of SSM tools in inferring landmarks for both control and pathology subjects when trained using only the morphology of one of these groups.  
Testing samples of controls and pathology for split-1 and split-2, respectively, were randomly sampled without replacement using a 25\%/75\% test/train split to validate landmark inference on held-out samples from the same group considered to build the shape model of each split.
The training data of each split were fed separately to SSM tools to generate point correspondences for the two splits. Point correspondences for the testing samples were estimated as in LAA. 

\noindent\textbf{Humerus.} Two random splits, split-1 and split-2, were defined similarly to the scapula dataset.

\noindent\textbf{Femur.} The data split for the femurs was random, and the split-1 and split-2 (similar to the scapula dataset) were used for evaluation. Split-1 alone was used for lesion screening.

%-------------------------------------------------------------------------------------
\subsection{Evaluation results}
%-------------------------------------------------------------------------------------
Evaluation of shape models is performed using quantitative and qualitative metrics detailed in Section \ref{sec:ssm-eval}. % compactness, generalization, specificity, dominant modes of variation, and clustering analysis (based on the anatomy). 
The compactness and specificity metrics are obtained using the training data. The generalization metric is computed as the ability of the shape model to represent held-out (i.e., testing) samples.

\begin{figure}%[ptb!]
    \centerline{\includegraphics[width = 1\linewidth]{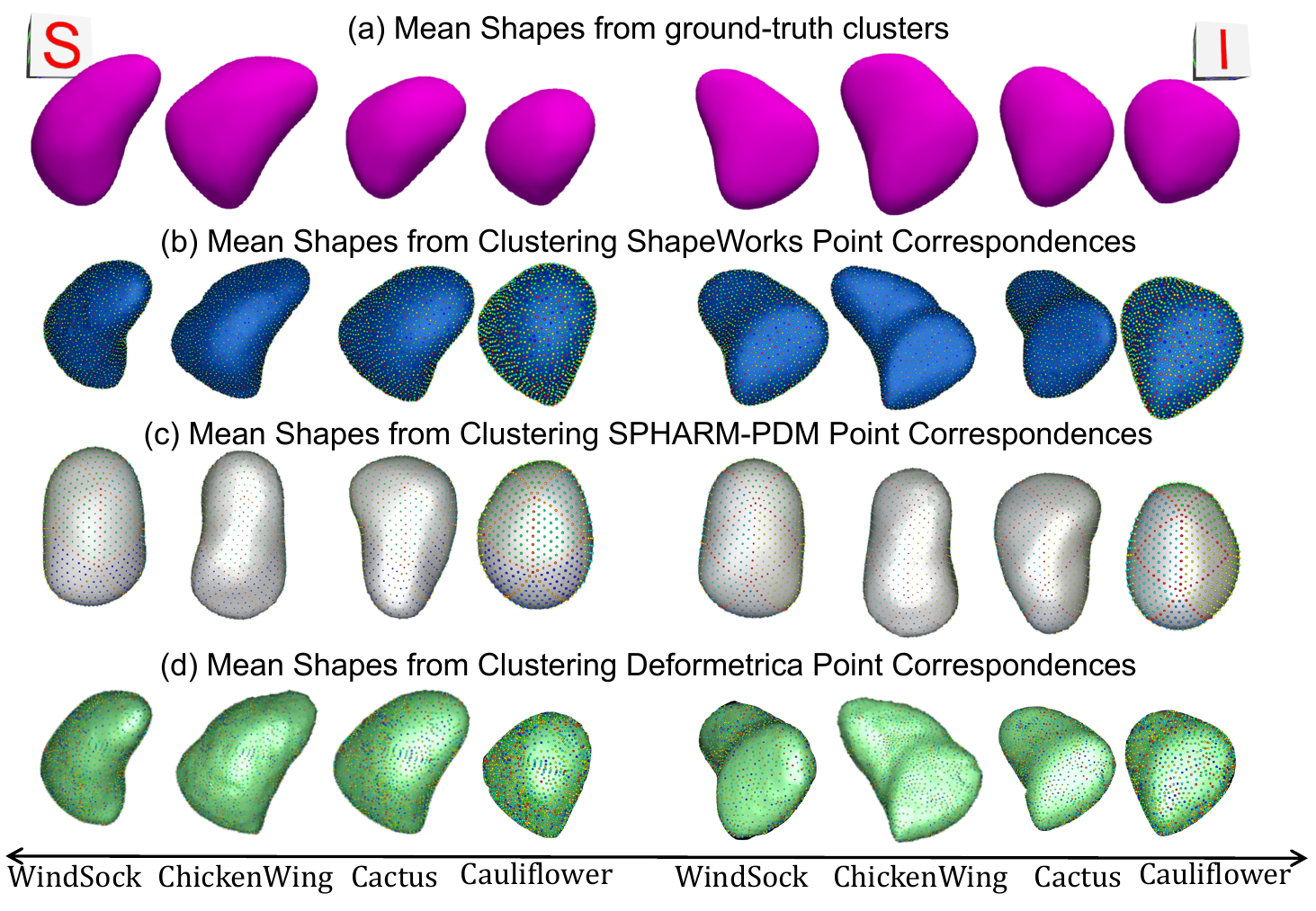}}
    \caption{ Superior (S) and inferior (I) views of mean shapes from (a) ground-truth clusters, and k-means clustering of correspondences from (b) ShapeWorks \cite{cates2017shapeworks}, (c) SPHARM-PDM \cite{styner2006framework}, and (d) Deformetrica \cite{durrleman2014morphometry}.  Cluster centers from ShapeWorks and Deformetrica models closely align with the ground-truth cluster centers.} % Adapted with permission from \cite{goparaju2018evaluation}.}%
    \label{LAAClusterMeanShapes}
\end{figure}

% \subsubsection{Anatomical measurements inference -- LAA}
\subsubsection{LAA shape models}

\figurename~\ref{evaluation_metrics_all}(a) shows the quantitative metrics (compactness, generalization, and specificity) of LAA shape models trained using the two random splits. 
ShapeWorks consistently produced a compact model compared to SPHARM-PDM and Deformetrica (first column).  %, as shown in first column of \figurename~\ref{evaluation_metrics_all}(a). 
ShapeWorks generalized better compared to SPHARM-PDM and Deformetrica in estimating the shape representation of unseen samples (second column). % (second column of \figurename~\ref{evaluation_metrics_all}(a)). 
Deformetrica outperformed ShapeWorks and SPHARM-PDM in the specificity measure (third column). %, as shown in the third column of \figurename~\ref{evaluation_metrics_all}(a). 
\figurename~\ref{modes_of_variation_all}(a) demonstrates the first two dominant modes of variation from the entire dataset without any splits. ShapeWorks and Deformetrica models were able to discover clinically relevant modes of variation in the data, which are elongation of the appendage and ostia size. 
The SPHARM-PDM model could discover neither the representative shape nor the dominant modes of variation correctly. % (\figurename~\ref{modes_of_variation_all}(a)).  
The clustering analysis was performed on all the samples without any training and testing splits. Four clusters were identified in the data using the elbow method. %as shown in figure~\ref{elbow}. 
The ability of shape models to discover the natural clusters was assessed as follows: The signed distance transform (DT) images were clustered using K-means, and the mean shape from each cluster was obtained to serve as a baseline. 
The ground-truth cluster labels for all the input shapes were manually annotated and reviewed by a clinical expert. The point correspondences from each shape model were clustered, and the cluster centers discovered from each tool were qualitatively compared to the mean shapes of the ground-truth clusters. 
The results illustrated in \figurename~\ref{LAAClusterMeanShapes} suggest that ShapeWorks and Deformetrica were able to discover the natural clusters in the data.

% \subsubsection{Anatomical landmarks estimation - scapula}
\subsubsection{Scapula shape models}

\begin{figure*}%[ptb!]
    \centerline{\includegraphics[width = 0.83\linewidth]{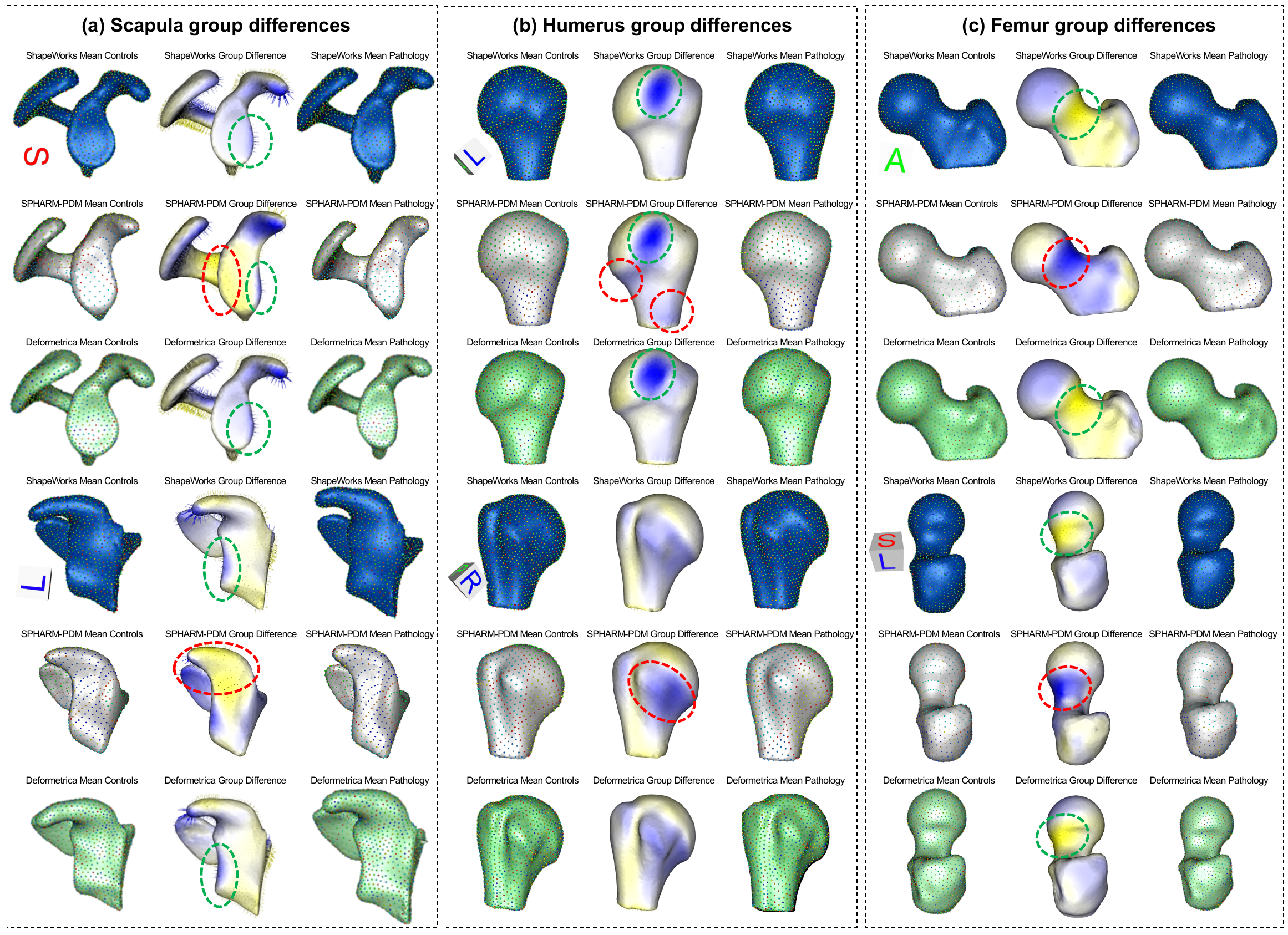}}
    \caption{Mean controls, mean pathology, and group difference for (a) scapula, (b) humerus, and (c) femur anatomies. (a) Superior (S) and left (L) views are shown for scapula. (b) Left (L) and right (R) views are shown for humerus. (c) Anterior (A) and superior-left (S-L) views are shown for femur.}%
    \label{group_differences_all}
\end{figure*}

\figurename~\ref{evaluation_metrics_all}(b) shows the quantitative metrics of scapula shape models. % trained using the two random splits. 
ShapeWorks consistently produced a compact model compared to SPHARM-PDM and Deformetrica for the two random splits. The generalization of Deformetrica and ShapeWorks was comparable in modeling unseen samples. 
Deformetrica specificity was better than that of ShapeWorks and SPHARM-PDM in split-1. The Deformetrica and ShapeWorks specificity measure were comparable in split-2. 
However, \figurename~\ref{evaluation_metrics_all}(b) shows that SPHARM-PDM could not generalize well, and the samples generated by the shape model were not representative of the shape population in both splits.
\figurename~\ref{modes_of_variation_all}(b) shows the dominant modes of variation in the entire dataset for controls and pathology. ShapeWorks and Deformetrica were able to discover the clinically relevant mode of variation, which is the variation of the glenoid size due to an anterior glenoid defect in the pathology subjects in the population. 
However, SPHARM-PDM could neither produce a representative shape of the population nor encode a clinically relevant mode of variation. 
%
% The group differences between controls and pathology were visualized using ShapeWorksView \cite{cates2017shapeworks}. \cite{cates8hypothesis}
%
Furthermore, ShapeWorks and Deformetrica models were able to capture the clinically relevant group differences between the controls and pathology population (see \figurename~\ref{group_differences_all}(a)).

% \subsubsection{Anatomical landmarks estimation -- humerus}
\subsubsection{Humerus shape models}

\figurename~\ref{evaluation_metrics_all}(c) shows the quantitative metrics of humerus shape models trained using the two random splits. 
SPHARM-PDM produced a compact model in split-1, and ShapeWorks produced a compact model in split-2. %, as shown in \figurename~\ref{evaluation_metrics_all}(c). 
Shapeworks outperformed both Deformetrica and SPHARM-PDM in generalizing well on held-out shapes and generating plausible and realistic shapes. 
The dominant modes of variation in the entire dataset were analyzed from the entire dataset consisting of controls and pathology. The first dominant mode of variation, which is the characterization Hill-Sachs lesion, as illustrated in \figurename~\ref{modes_of_variation_all}(c), was identified correctly by all the models. 
%
% The group differences between controls and pathology were visualized using ShapeWorksView \cite{cates2017shapeworks}. 
%
Moreover, all the models were able to capture the clinically relevant group differences between the controls and pathology populations (see \figurename~\ref{group_differences_all}(b)). Nonetheless, models from SPHARM-PDM encoded differences that are not aligned with the underlying morphological characteristics of the Hill-Sachs lesion.

% \subsubsection{Lesion screening -- femur and humerus}
\subsubsection{Femur shape models}

\figurename~\ref{evaluation_metrics_all}(d) shows the quantitative metrics of femur shape models trained using the two random splits. 
SPHARM-PDM consistently produced a compact model compared to ShapeWorks and Deformetrica for the two random splits. %, as shown in \figurename~\ref{evaluation_metrics_all}(d). 
The generalization of ShapeWorks was better than that of Deformetrica and SPHARM-PDM in modeling unseen samples. 
The specificity of ShapeWorks was better than that of Deformetrica and SPHARM in both splits. 
However, \figurename~\ref{evaluation_metrics_all}(d) shows that SPHARM-PDM could not generalize well, and the samples generated by the shape model were not representative of the shape population in both splits.
The dominant modes of variation in the femur data were analyzed from the entire dataset consisting of controls and pathology. 
ShapeWorks and Deformetrica were able to discover the clinically relevant mode of variation, which is the extra bone growth in the femoral head (see \figurename~\ref{modes_of_variation_all}(c)). 
However, SPHARM-PDM could not encode the clinically relevant mode of variation.
%
% The group differences between controls and pathology were visualized using ShapeWorksView \cite{cates2017shapeworks}. 
ShapeWorks and Deformetrica models were also able to capture the clinically relevant group differences between the controls and pathology population (see \figurename~\ref{group_differences_all}(c)).

\begin{figure*}%[ptb!]
    \centerline{\includegraphics[width = \linewidth]{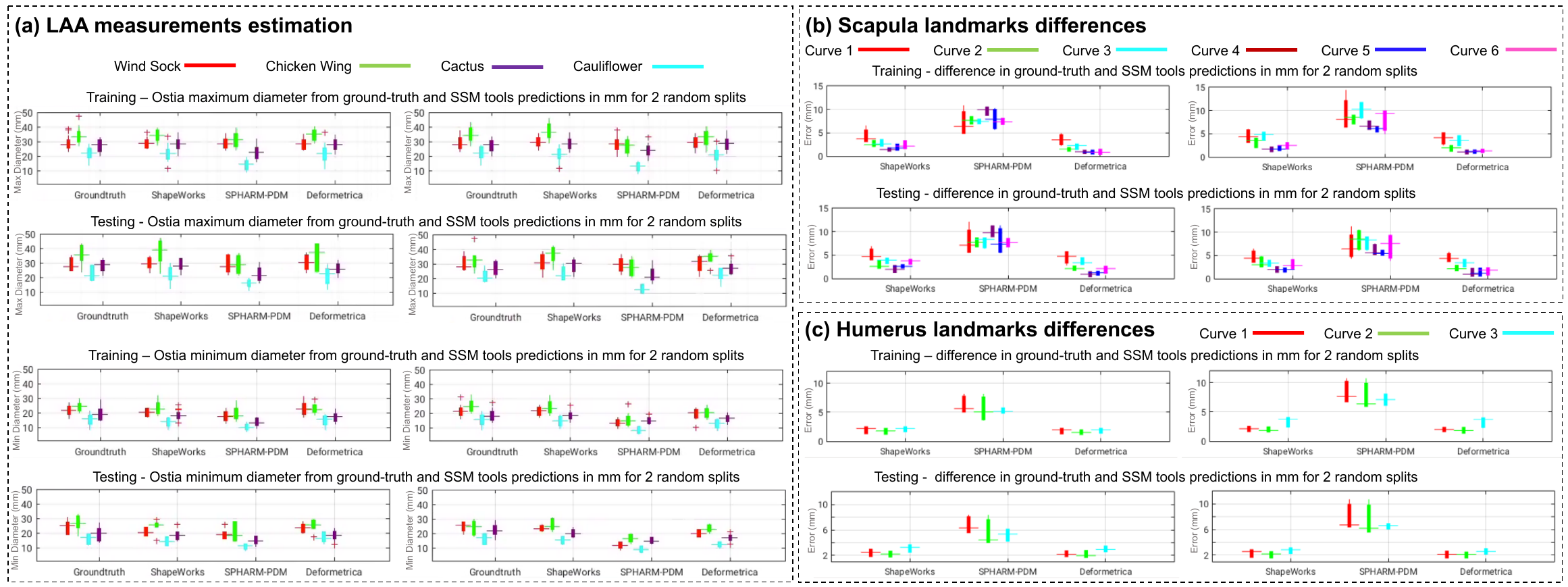}}
    \caption{Validation results of the (a) LAA ostia maximum and minimum diameter measurements, (b) scapula landmark differences, and (c) humerus landmark differences, from ground-truth and predictions of SSM tools in mm for the two random splits. }
    \label{measurements_landmarks_all}
\end{figure*}

%-------------------------------------------------------------------------------------
\subsection{Validation results}
%-------------------------------------------------------------------------------------
The validation is conducted by comparing the ground-truth information with the predictions of the SSM tools. % in the context of clinical applications. 

\subsubsection{Anatomical measurements inference -- LAA}

SSM tools were validated based on the accuracy of the LAA ostia measurement predictions. The ground-truth measurements were obtained, as shown in \figurename~\ref{LAA}, where the landmarks on the LAA ostium were used to compute the ground-truth measurements of the LAA maximum and minimum diameters by fitting an ellipse to each LAA ostium.
%
% SSM tool predictions were obtained as follows:
% The cluster mean shapes were computed using the point correspondences from the training data.
%
% The ostium contour was manually marked using ParaView \cite{ayachit2015paraview} for each cluster mean shape and warped from the mean space to the subject space of each training and testing sample using a thin plate spline warp \cite{bookstein1989principal}. 
%
% The subject-specific ostium measurements were obtained by fitting an ellipse for the validation purposes. 
%
\figurename~\ref{measurements_landmarks_all}(a) shows the ground-truth measurements and SSM tool predictions for the LAA ostia maximum and minimum diameters for the training and testing samples. 
ShapeWorks and Deformetrica models predictions were closely aligned to the ground-truth compared to predictions from SPHARM-PDM models.
Statistical tests showed the equivalence of the predicted and ground measurements, based on Euclidean distances, for ShapeWorks and Deformetrica in split-1 for the maximum diameter (p = 0.569 and 0.210, respectively), and Deformetrica in split-2 (p = 0.436).  When combining the splits with clusters, we found the equivalence for ShapeWorks, SPHARM-PDM, and Deformetrica (maximum diameter) for split-1-cluster-1, split-2-cluster-1, and split-2-cluster-2.  In addition, when using Deformetrica, we found the equivalence (for maximum diameter) in all splits and cluster combinations except for split-1-cluster-4 (p = 0.037).

\subsubsection{Anatomical landmarks estimation -- scapula}

% SSM tools are validated based on the accuracy of the scapulae landmarks inference. 
%
% SSM-based landmarks inference can be summarized as follows:
%
% The ground-truth landmarks of the training data were used to compute the average training landmarks. Then, the training mean shape was computed using the point correspondences from the training data.
%
% The mean training shape and the average training landmarks were warped from the mean space to the subject space using a thin plate spline \cite{bookstein1989principal} warp for each subject to obtain subject-specific landmarks. 
%
% The details of obtaining the average training landmarks from different SSM tools are discussed in section ~\ref{methods_landmarks}. Next, 
%
% The subject-specific warping of landmarks was performed on training and testing samples to compute the accuracy of landmark predictions for training and testing samples. 
%
\figurename~\ref{measurements_landmarks_all}(b) shows the Euclidean distance between the ground-truth landmarks and landmarks inference from each SSM tool (average of the cumulative distances for the points/landmarks on each curve) for the six anatomical curves of scapula. We found smaller errors in the case of curves 4, 5, and 6 in the two random splits. The performance of the Deformetrica and ShapeWorks models is comparable and better than that of the SPHARM-PDM model. % in scapula landmarks inference.
The measurement of the glenoid radius can be computed from the landmarks of curve 4. The glenoid radius was obtained from the ground-truth landmarks and inferred landmarks of the SSM tools.
In split-1, statistical tests showed the equivalence of the predicted and ground-truth measurements, based on the distances in Euclidean space, for the glenoid radius in ShapeWorks (p = 0.07 for no template for initialization and 0.09 for the mean template for initialization), for the distance between the apex of the coracoid process and anterolateral corner (ALC) of the acromion as well as the distance between apex of the coracoid process and the posterolateral corner (PLC) of the acromion in Deformetrica (p = 0.112 and 0.209, respectively, for the raw measurements; p = 0.416 and 0.140, respectively, for the ellipse atlas; p = 0.355 and 0.168, respectively, for the sphere atlas; p = 0.149 and 0.285 for the medoid atlas).

\subsubsection{Anatomical landmarks estimation -- humerus}

% SSM tools were validated based on the accuracy of the humerus landmarks inference. 
%
% To obtain the SSM tool predictions,
% the ground-truth landmarks of the training data were used to compute the average training landmarks. 
%
% The training mean shape was computed using the shape vectors from the training data.
%
% The mean training shape and the average training landmarks were warped from the mean space to the subject space using a thin plate spline \cite{bookstein1989principal} warp for each subject to obtain subject-specific landmarks. 
%
% The details of obtaining the average training landmarks from different SSM tools are discussed in section ~\ref{methods_landmarks}. The subject-specific warping of landmarks was performed on training and testing samples to get the accuracy of landmark predictions in the training and testing phases.
%
The landmarks inference of Deformetrica and ShapeWorks was better than that of SPHARM-PDM, resulting in fewer errors (see \figurename~\ref{measurements_landmarks_all}(c)). Curve 2 had fewer errors compared to curves 1 and 3 in both the random test data predictions.
The measurement humerus radius can be computed from the landmarks of curve 2. The humerus radius was obtained from the ground-truth landmarks and the inferred landmarks of the SSM tools.
Statistical tests showed the equivalence of the predicted and ground truth measurements, based on Euclidean distances, for Deformetrica in split-1 for the humerus head radius (p = 0.09). 

\begin{figure*}
    \centering
    \includegraphics[width=\linewidth]{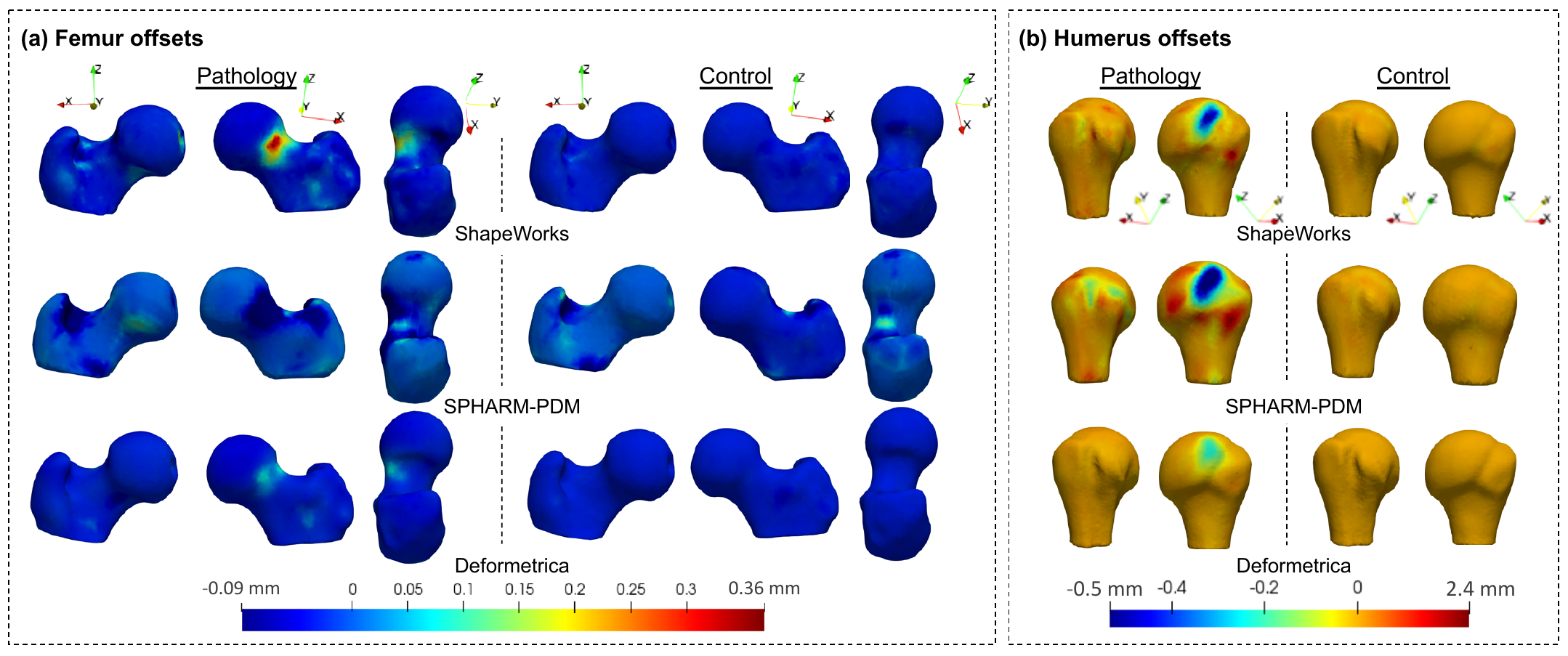}
    \caption{The group differences between the reconstructed samples and reconstructed samples with offsets. (a) Femur group differences for pathology (left) and controls (right). (b) Humerus group differences for pathology (left) and controls (right).}
    \label{offsets_all}
\end{figure*}

\begin{figure*}%[ptb!]
    \centering
    \includegraphics[width = 0.9\linewidth]{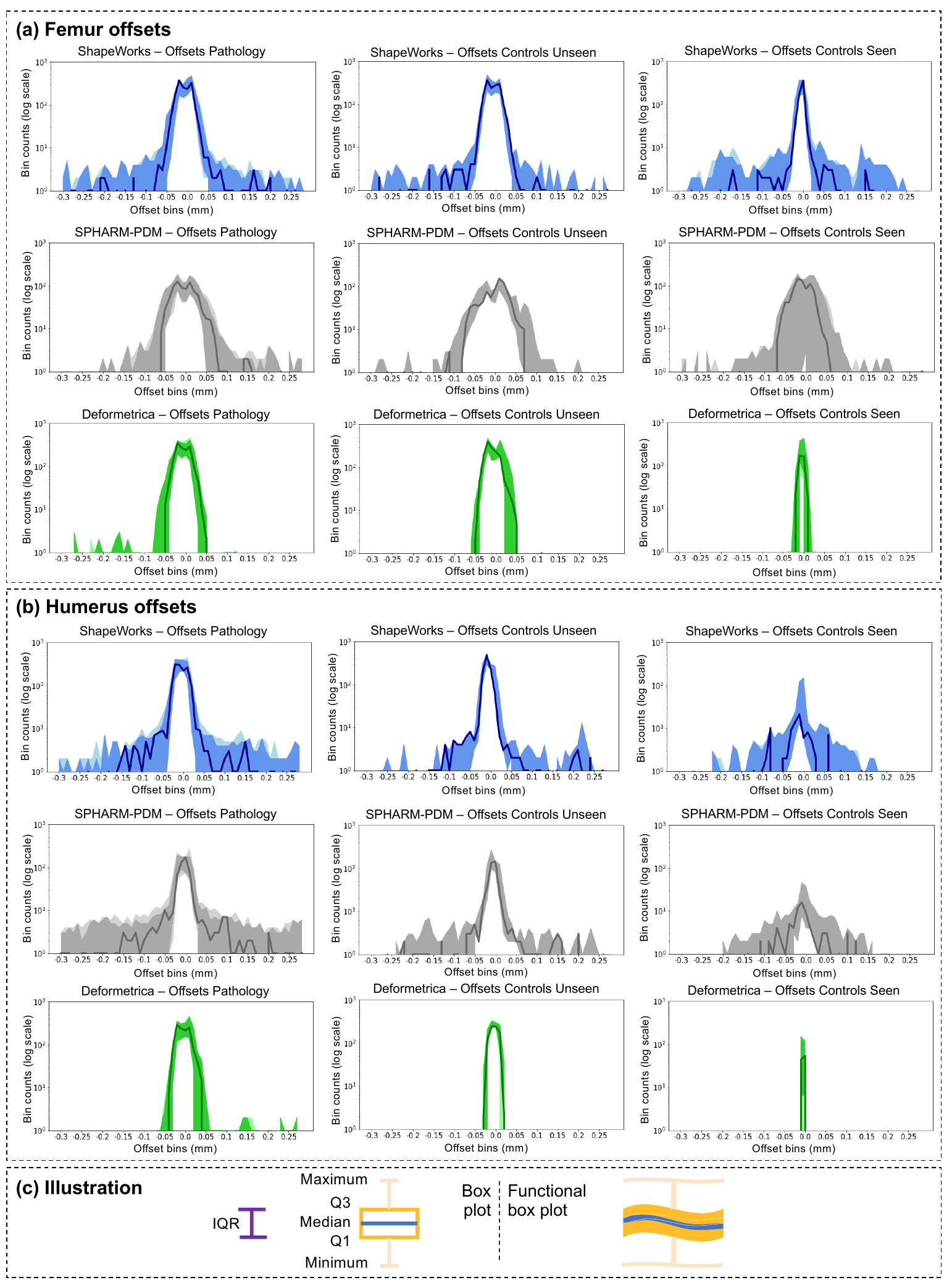}
    \caption{Functional box plots of histograms of the offset values identified by slack variables optimization process for pathology, held-out controls, and training control samples of (a) femur and (b) humerus. (c) Illustration of functional box plot \cite{sun2011functional}. The IQR region, the median curve (blue), and the non-outlying region are demonstrated as descriptive statistics.}%
    \label{offsets_plots_all}
\end{figure*}

\subsubsection{Lesion screening --  femur and humerus}

SSM tools were validated based on the lesion identification and the accuracy of the classification of the pathology. 
The lesion identification is qualitative because the ground-truth lesion is unavailable for the participants with pathology. 
The accuracy of classification of the pathology from shape models was obtained to quantify the performance.
%

% The trained shape model from controls was obtained from each SSM tool. Each pathology sample was projected and reconstructed from the shape model subspace to obtain the closest control sample . 
%
% The control subjects (seen during the training process and held-out samples) were also projected and reconstructed from the shape model subspace. The differences between the pathology sample and the reconstruction can potentially signify the lesion. 
%
% Due to the possibility of false positives in this process, the correspondence points on the lesion were down-weighted, and the closest control was obtained using an optimization process with the slack variables approach. 
%

\vspace{0.05in}
\noindent\textbf{Lesion identification:} The slack variable optimization (Algorithm 1) resulted in the identification of the closest control to the pathology and captured the lesion in the slack variables or offsets in the normal direction of each correspondence point. 
%
% The difference between the reconstruction and reconstruction with the offsets in the normal direction of correspondences assisted in lesion identification. 
%
The differences between the reconstruction and reconstruction with the offsets in the normal direction were visualized groupwise for all the control and pathology samples. 
The offsets for the controls did not signify a lesion, whereas the offsets for the pathology signified a lesion. 
For femurs, the lesion was correctly identified in the case pathological group differences by ShapeWorks and Deformetrica models (see \figurename~\ref{offsets_all}(a).
The offsets from the optimization process were visualized using functional box plots \cite{sun2011functional}. 
The interquartile range (IQR) is indicated as a band in a color. The median is a curve within the IQR, the region outside the IQR is another color, and the nonoutlying region is a band in the plots (see \figurename~\ref{offsets_plots_all}(c)).
The offset values between -0.005 and 0.005 were set to 0 to visualize the offsets trend (see \figurename~\ref{offsets_plots_all}(a)). 
The offsets from ShapeWorks and Deformetrica for the pathology samples were mostly positive, which indicates a lesion (extra bone growth). 
The band of offsets for the control subjects (seen, held-out, or unseen) was narrower compared to that of the pathology subjects. 
For humeri, the lesion was correctly identified in the case of pathological group differences by all the models (see \figurename~\ref{offsets_all}(b)). 
The SPHARM-PDM model captured false positives in the pathology differences compared to ShapeWorks and Deformetrica models (see \figurename~\ref{offsets_all}(b)). 
%
% The offsets from the optimization process were visualized using the functional box plots \cite{sun2011functional}. The offset values between -0.005 and 0.005 were set to 0 to visualize the offsets trend (see \figurename~\ref{offsets_plots_all}(b)). 
The offsets from all the models for the pathology samples were mostly negative, indicating a lesion (bone loss), see \figurename~\ref{offsets_plots_all}(b). 
The band of offsets for the control subjects (seen, held-out, or unseen) was narrower compared to that of the pathology subjects.

\vspace{0.05in}
\noindent\textbf{Pathology classification:} The offsets obtained from the optimization process were fed to a multilayer perceptron with labels 0 and 1 indicating control and pathology, respectively. The dataset was randomly split into training and testing sets. 
The best set of hyperparameters (activation, hidden layers, number of units in each hidden layer, regularization, and solver) was found using three-fold cross validation on the training data of each SSM tool independently.
The model was then trained using the training data with the best set of hyperparameters and used for the classification of pathology on the test data. The random train-test split, hyperparameter tuning, and testing were performed for several iterations to obtain the average predictions from the trained models. 
Classification performance metrics %such as average accuracy, standard deviations of accuracies, average F1-score, standard deviation of F1-score, average area under the curve, and standard deviation of area of the curve 
were obtained from the trained models on the training and testing data (see Table \ref{pathology_classification}). 
In the case of the femur, the performance of ShapeWorks and Deformetrica trained models was comparable (see Table \ref{pathology_classification}(a)). The standard deviation of metrics was low for ShapeWorks, indicating the minimal deviation of the results for different train-test splits. SPHARM-PDM results were inferior in the classification of pathology.
In the case of the humerus, the performance of ShapeWorks trained model was better than that of Deformetrica and SPHARM-PDM (see Table \ref{pathology_classification}(b)). SPHARM-PDM model performance was comparable to that of ShapeWorks. 
The standard deviation of metrics was relatively higher for Deformetrica-trained models.

\begin{table*}
\caption{Pathology classification training performance from SSM tools}
\label{pathology_classification}
\vspace{\baselineskip}
\begin{center}
\resizebox{\linewidth}{!}{%
\begin{tabular}{|c|c|c|c|c|c|c|}
\hline 
\multirow{2}{*}{} & \multicolumn{6}{c|}{\textbf{(a) Femur pathology classification}}\tabularnewline
\cline{2-7} \cline{3-7} \cline{4-7} \cline{5-7} \cline{6-7} \cline{7-7} 
 & \multicolumn{3}{c|}{\textbf{Training performance}} & \multicolumn{3}{c|}{\textbf{Testing performance}}\tabularnewline
\hline 
 \backslashbox{\textbf{Metrics}}{\textbf{SSM Tools}}  & \textbf{ShapeWorks} & \textbf{SPHARM-PDM} & \textbf{Deformetrica} & \textbf{ShapeWorks} & \textbf{SPHARM-PDM} & \textbf{Deformetrica}\tabularnewline
\hline 
\textbf{Accuracy ($\mu \pm \sigma$)\%} &  \textbf{97.851} $\pm$ \textbf{0.025}  & 77.851 $\pm$  0.088 & 93.465 $\pm$ 0.03  &  83.167 $\pm$ 0.078 & 55.833 $\pm$ 0.085 & \textbf{83.833} $\pm$ \textbf{0.083}\tabularnewline
\hline 
\textbf{F1 Score ($\mu \pm \sigma$)\%} &   \textbf{96.977} $\pm$ \textbf{0.033} & 60.681 $\pm$ 0.198 & 90.634 $\pm$ 0.043  & \textbf{81.125} $\pm$ \textbf{0.108} & 39.455 $\pm$ 0.203 & 80.89 $\pm$ 0.128 \tabularnewline
\hline 
\textbf{AUC ($\mu \pm \sigma$)} &  \textbf{0.697} $\pm$ \textbf{0.109} & 0.378 $\pm$ 0.145 & 0.595 $\pm$ 0.278  &  \textbf{0.613} $\pm$ \textbf{0.166} & 0.52 $\pm$ 0.298 & 0.57 $\pm$ 0.135  \tabularnewline
\hline 
\multirow{2}{*}{} & \multicolumn{6}{c|}{\textbf{(b) Humerus pathology classification}}\tabularnewline
\cline{2-7} \cline{3-7} \cline{4-7} \cline{5-7} \cline{6-7} \cline{7-7} 
 & \multicolumn{3}{c|}{\textbf{Training performance}} & \multicolumn{3}{c|}{\textbf{Testing performance}}\tabularnewline
\hline 
 \backslashbox{\textbf{Metrics}}{\textbf{SSM Tools}} & \textbf{ShapeWorks} & \textbf{SPHARM-PDM} & \textbf{Deformetrica} & \textbf{ShapeWorks} & \textbf{SPHARM-PDM} & \textbf{Deformetrica}\tabularnewline
\hline 
\textbf{Accuracy ($\mu \pm \sigma$)\%} &  \textbf{98.73} $\pm$ \textbf{0.012} & 97.143 $\pm$ 0.019 & 93.651 $\pm$ 0.119  &  \textbf{96} $\pm$ \textbf{0.033} & 95.333 $\pm$ 0.027 & 90.667 $\pm$ 0.122 \tabularnewline
\hline 
\textbf{F1 Score ($\mu \pm \sigma$)\%} & \textbf{98.912} $\pm$ \textbf{0.01} & 97.619 $\pm$ 0.015 & 92.824 $\pm$ 0.137  &   \textbf{96.157} $\pm$ \textbf{0.029} & 95.108 $\pm$ 0.031 & 87.46 $\pm$ 0.188 \tabularnewline
\hline 
\textbf{AUC ($\mu \pm \sigma$)} &  0.829 $\pm$ 0.134 & \textbf{0.853} $\pm$ \textbf{0.07} & 0.796 $\pm$ 0.262  & \textbf{0.88} $\pm$ \textbf{0.102} & 0.822 $\pm$ 0.13 & 0.762 $\pm$ 0.349 \tabularnewline
\hline 
\end{tabular}
}
\end{center}
\end{table*}

%-------------------------------------------------------------------------------------
\section{Discussion}
%-------------------------------------------------------------------------------------

ShapeWorks produced shape models with consistent quantitative and qualitative performances in most of the experiments detailed in the results section. This consistency can be attributed to the underlying groupwise correspondence-based approach.
For evaluation metrics, ShapeWorks resulted in compact models of the LAA, scapula, and humerus anatomies (see \figurename~\ref{evaluation_metrics_all}). 
ShapeWorks models generalized well for the LAA, scapula, humerus, and femur anatomies, and consistently generated plausible shapes of the scapula, humerus, and femur anatomies.
ShapeWorks models were able to discover clinically relevant modes of variation, including the group differences for all the studied anatomies, and the natural clusters in LAA (see \figurename~\ref{LAAClusterMeanShapes}) and its validation outcomes were closely aligned to the ground-truth.

\begin{figure}[ptb!]
    \centerline{\includegraphics[width = 1\linewidth]{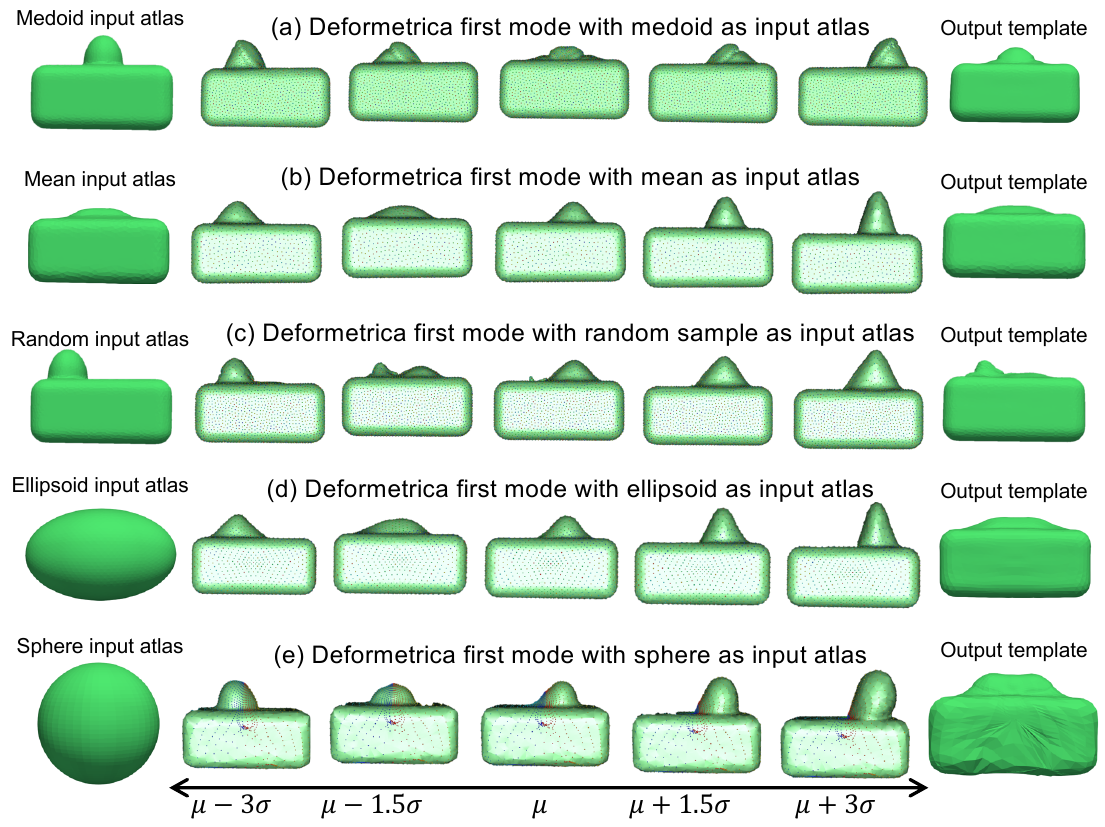}}
    \caption{ Box bump mode of variation of Deformetrica with the input atlas as (a) the medoid, (b) the mean, (c) a random shape, (d) an ellipsoid, and (e) a sphere, producing different shape statistics.  }%
    \label{deform-box-bum-templates}
\end{figure}

\begin{figure}[ptb!]
    \centerline{\includegraphics[width = 1\linewidth]{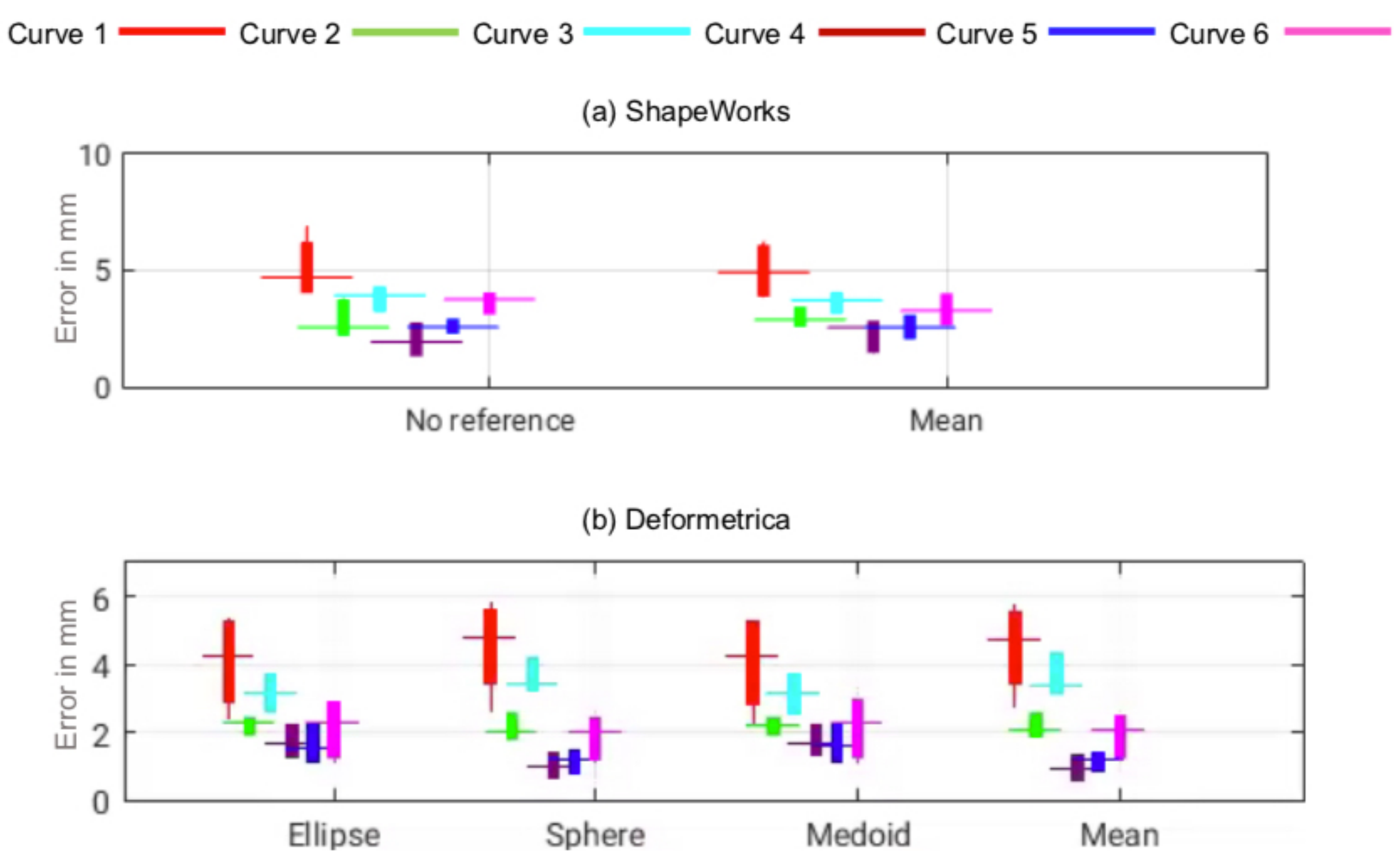}}
    \caption{ Quantitative assessment (scapula landmarks inference task) of SSM tools with different input atlases on the unseen samples. (a) ShapeWorks with no reference  and input initialization as a mean shape, (b) Deformetrica with input atlases as ellipse, sphere, medoid, and mean shapes.  }%
    \label{diff-templates}
\end{figure}

Deformetrica models were comparable to those of ShapeWorks in a few experiments due to the underlying groupwise deformation-based approach. 
However, Deformetrica results were not consistent throughout the experiments because of the impact of the input atlas that needs to serve as an initialization.
A qualitative assessment of the performance of Deformetrica models with different atlases was performed on an ensemble of 3D shapes of boxes with a moving bump.
The first mode of variation from the Deformetrica models with different input atlases (mean, medoid, random input, ellipsoid, and sphere) resulted in large variability in the first mode of variation (see \figurename~\ref{deform-box-bum-templates}). 
The ellipsoid and sphere atlases were scaled to match the input shapes. The variability displayed in the discovery of the moving bump informs the inconsistency in the Deformetrica models. 
When the medoid was provided as an input atlas to Deformetrica, the moving bump in the first mode of variation was closely aligned to the ground-truth. The modes of variation for the mean and ellipsoid input atlases were similar.
A quantitative assessment of the performance of Deformetrica models with different atlases was performed on scapula landmarks inference task.
The algorithm could not produce a good shape model when the input atlas was provided as an ellipse and sphere. Hence, the sphere and ellipsoid were deformed onto some subject as a preprocessing step. The deformed sphere and ellipsoid at an intermediate step of the deformation flow were then used as modified initial atlases.
The Euclidean distance between the ground-truth and predicted landmarks from the Deformetrica models with different input atlases (ellipsoid, sphere, medoid, and mean) resulted in different levels of errors (see \figurename~\ref{diff-templates} (b)). 
ShapeWorks does not need an input atlas to generate point correspondences. To analyze the performance of ShapeWorks with an input atlas, the point correspondences of the training data were initialized to the mean training shape. The Euclidean distance between the ground-truth and predicted landmarks with no reference and mean shape initialization was compared (see \figurename~\ref{diff-templates} (a)).

SPHARM-PDM models mostly displayed inferior results compared to those of Deformetrica and ShapeWorks in the evaluation and validation experiments.
This inferior performance can be attributed to the pairwise correspondence-based approach that does observe the entire cohort, where the correspondences from SPHARM-PDM are generated by mapping every input shape to a unit sphere. 
This spherical mapping can result in ambiguity in the mapping of the axes demonstrated in the LAA modes of variation (see \figurename~\ref{modes_of_variation_all}(a)). 
In the case of evaluation metrics, SPHARM-PDM could not produce compact models of the anatomies LAA, scapula, and humerus (see \figurename~\ref{evaluation_metrics_all}.
SPHARM-PDM models could not generalize adequately for the scapula, humerus, and femur anatomies, and could not generate plausible shapes for all the anatomies.
The SPHARM-PDM models could not consistently discover clinically relevant modes of variation, including the group differences, and were unable to discover natural clusters in LAA (see \figurename~\ref{LAAClusterMeanShapes}). 
SPHARM-PDM validation outcomes were rarely aligned to the ground-truth.

In summary, the SSM tools produced different levels of consistency in the evaluation and validation process, which indicates the need for such an assessment in real-world clinical applications. 
Based on the overall results from all the experiments, we can infer that the groupwise correspondence technique can potentially learn the population-level variability compared to the pairwise correspondence method.

%-------------------------------------------------------------------------------------
\section{Conclusion and future work}
%-------------------------------------------------------------------------------------
The main contribution of this work is a systematic evaluation and validation of open-source statistical shape modeling (SSM) tools in the context of clinical applications, an area in which there has been little work \cite{gollmer2014}.

%-------------------------------------------------------------------------------------
\subsection{Research contributions}
%-------------------------------------------------------------------------------------

In this paper, we have presented an evaluation and clinically driven validation framework to assess the performance of shape models from different open-source SSM tools. 
Quantifying the performance of shape models is a challenging task due to the lack of ground-truth correspondences.
This problem has been addressed by considering qualitative and quantitative metrics to determine the utility of shape models in clinical applications.
The evaluation of shape models is performed using quantitative metrics such as compactness, generalization, specificity \cite{davies20023d}, and qualitative metrics, including modes of variation and clustering analysis. 
The validation of shape models is performed based on the differences between ground-truth and SSM tool predictions of anatomical measurements and class.
The evaluation and validation framework is tested on representative real-world clinical applications such as implant design and selection, motion tracking, surgical planning, bone resection, and bone grafting. 
Different tools produced different levels of consistencies, which highlights the importance of such an assessment. ShapeWorks \cite{cates2017shapeworks} and Deformetrica \cite{durrleman2014morphometry} models displayed better results in the clinical applications compared to SPHARM-PDM \cite{styner2006framework} models due to the underlying groupwise approach for establishing shape correspondences.
Deformetrica models displayed inconsistencies in results due to the bias introduced by the input atlas used for initialization. SPHARM-PDM models were inferior in performance due to the underlying pairwise correspondence approach. 
The evaluation indicated that SPHARM-PDM models mostly were unable to produce compact models, generalize well to unseen shapes, and generate realistic shapes that retain the shape characteristics of the population under study. 
SPHARM-PDM models could not discover clinically relevant modes of variation and could not identify natural clusters in a morphology such as LAA, due to ambiguity in the mapping of the axes. 
The validation demonstrated that ShapeWorks and Deformetrica models were comparable in performance and outperformed SPHARM-PDM models. %, producing fewer errors.

%-------------------------------------------------------------------------------------
\subsection{Scientific impact}
%-------------------------------------------------------------------------------------

This research provides a direction to systematically assess different SSM tools available for clinical applications. 
The framework assists in selecting and deploying the right SSM tool to address a clinical need. 
The assessment of SSM tools can motivate further research and enhancement of the underlying optimization techniques involved in shape-modeling tools. 
Benchmarking the performance of shape models could motivate the development of a new class of shape-modeling tools and techniques, which could take the performance of SSM in real-world applications to another level. 
This study may also drive the development of a new set of tools to automate the end-to-end evaluation and validation of SSM tools, when given training and test data. 
The evaluation and validation framework proposed in this paper could easily be extended to other clinical situations or other classes of applications of SSM.

%-------------------------------------------------------------------------------------
\subsection{Limitations and future work}
%-------------------------------------------------------------------------------------

This research is confined to three open-source, widely used, state-of-the-art SSM tools applicable for general anatomies. However, the framework can be adapted to other SSM tools that work on general purpose anatomies or SSM tools that are tailored to specific anatomies. 
The performance results of the SSM tools discussed in this paper cannot be baselined for all the clinical applications or other clinical scenarios. 
The results from SSM tools can vary based on the various steps followed in the shape-modeling process, such as training data collection, data preprocessing, and parameter tuning for the shape models. High-quality training data can help improve the shape-modeling process.
In the future, this study can be extended to other publicly available tools and clinical applications to benchmark SSM tools in different scenarios and
to provide a blueprint for the development of computational methods, tools, and techniques for shape modeling.

%Matthijs Jacxsens, Ibolya Csecs, Alan Morris, and Evgueni Kholmovski for assisting in the manual annotation of anatomical landmarks and image acquisition.

\section*{Acknowledgements}

This work was supported by Coherex Medical and the National Institutes of Health under grant numbers NIBIB-U24EB029011, NIAMS-R01AR076120, NHLBI-R01HL135568, and NIGMS-P41GM103545. 
The content is solely the responsibility of the authors and does not necessarily represent the official views of the National Institutes of Health.
This work was also partly funded by the European Research Council under grant number 678304, European Union’s Horizon 2020 research and innovation program under grant number 666992, and program Investissements $d^\prime$avenir under grant number ANR-10-IAIHU-06.
The authors would like to thank the Division of Cardiovascular Medicine (data were collected under Nassir Marrouche, MD, oversight and currently managed by Brent Wilson, MD, PhD) and the Orthopaedic Research Laboratory (ORL) at the University of Utah for providing the MRI/CT scans and the corresponding segmentations of the left atrium appendage, femur, humerus and scapula, with a special thanks to Evgueni Kholmovski for assisting in MRI image acquisition. Authors also acknowledge Christine Pickett and Riddhish Bhalodia for reviewing the manuscript.

%%Harvard
\bibliographystyle{Style/model2-names.bst}\biboptions{authoryear}
\bibliography{refs}
\end{document}